\pdfoutput=1

\documentclass[11pt]{article}

\usepackage[preprint]{acl}

\usepackage{times}
\usepackage{latexsym}

\usepackage{booktabs}
\usepackage{multirow}
\usepackage{hhline}
\usepackage{array}
\usepackage{tabularx}  

\usepackage[T1]{fontenc}

\usepackage[utf8]{inputenc}

\usepackage{microtype}

\usepackage{inconsolata}

\usepackage{graphicx}
\usepackage{subcaption}
\usepackage{amsmath}
\usepackage{amsfonts}
\usepackage{algorithm}
\usepackage{algpseudocode}
\usepackage{hyperref}
\usepackage{booktabs}
\usepackage{multicol}
\usepackage{amssymb}
\usepackage{pifont}
\newcommand{\cmark}{\ding{51}}%
\newcommand{\xmark}{\ding{55}}%

\NewDocumentCommand{\heng}
{ mO{} }{\textcolor{red}{\textsuperscript{\textit{Heng}}\textsf{\textbf{\small[#1]}}}}

\newif\iftodo

\title{
That’s Deprecated! Understanding, Detecting, and Steering Knowledge Conflicts in Language Models for Code Generation
}

\author{Jaesung Bae\thanks{Equal contribution. Ordered in alphabet order.}, Cameron Churchwell\footnotemark[1], Mitchell Hermon\footnotemark[1], Tsun-An Hsieh\footnotemark[1], \\ \textbf{Jocelyn Xu\footnotemark[1], Yekaterina Yegorova\footnotemark[1], Mark Hasegawa-Johnson, Heng Ji}\\
University of Illinois Urbana-Champaign \\
\texttt{\{jb82, cc178, mhermon2, tsunanh2, yuex7, yay2, jhasegaw, hengji\}@illinois.edu}
}

\begin{document}
\maketitle

\begin{abstract}
This paper investigates how large language models (LLMs) behave when faced with discrepancies between their parametric knowledge and conflicting information contained in a prompt. Building on prior question-answering (QA) research, we extend the investigation of knowledge conflicts to the realm of code generation. We propose a domain-agnostic framework for constructing and interpreting such conflicts, along with a novel evaluation method and dataset tailored to code conflict scenarios. Our experiments indicate that sufficiently large LLMs encode the notion of a knowledge conflict in their parameters, enabling us to detect knowledge conflicts with up to \textbf{80.65\%} accuracy. Building on these insights, we show that activation-level steering can achieve up to a \textbf{12.6\%} improvement in steering success over a random baseline. However, effectiveness depends critically on balancing model size, task domain, and steering direction. The experiment code and data will be made publicly available after acceptance.

\end{abstract}

\section{Introduction}

Modern large language models (LLMs) face an inherent tension between their parametric knowledge (PK), encoded during training, and the potential conflicting knowledge (CK) they receive from user prompts. While this dual knowledge architecture enables adaptability across tasks ranging from natural language understanding to code generation, it creates a fundamental challenge: how can models reconcile new, potentially contradictory contextual information with their existing knowledge? This challenge becomes particularly salient in cases of \textbf{context-memory conflicts}, where contextual information directly contradicts a model's PK \cite{xuKnowledgeConflictsLLMs2024, suConflictBankBenchmarkEvaluating2024, qian2024merge, xie2023adaptive, tighidet-etal-2024-probing}. For instance, consider a Python library function that was updated or deprecated after the model’s training concluded. If the user prompt supplies the new information, a conflict arises between the model’s outdated parametric knowledge and the prompt’s updated details.

Understanding the mechanisms underlying LLM behavior in response to these conflicts is a crucial step toward developing models that can effectively identify, isolate, and navigate knowledge conflicts—key capabilities essential for reliable LLM deployment \cite{wangResolvingKnowledgeConflicts2024a}. Previous studies investigating knowledge conflicts \cite{wangResolvingKnowledgeConflicts2024a, longpreEntityBasedKnowledgeConflicts2021, richknowledgesources} have primarily focused on QA tasks, where responses typically consist of a single word or a short sequence of text.

To expand beyond the relatively concise nature of QA tasks, we examine the code generation domain—a crucial and growing application area for LLMs, where context-memory conflicts are both frequent and impactful. Code generation often entails multiple steps of reasoning, longer responses, and domain-specific knowledge, making it a rich setting to investigate how LLMs handle contradictory information. We introduce a framework to conceptualize context-memory conflicts and provide instantiations of this framework for both QA and code generation tasks. Then, we address three key research questions for both tasks: 1) \textit{How are knowledge conflicts treated in LLMs?}, 2) \textit{Can knowledge conflicts be detected?}, and 3) \textit{Can we steer the response of LLMs?}

To answer these research questions, we first study how models handle these conflicts, identifying patterns in response proportions and attention maps of LLMs across different tasks, model sizes, and statement types. Our experiments reveal that models tend to rely more on PK when tasks are easier, model sizes are larger, and additional endorsement statements are provided.
Second, we adopt probing techniques \cite{conneau-etal-2018-cram, belinkov2022probing, cho2023prompt, tighidet-etal-2024-probing} to detect which knowledge the LLMs rely on more. Specifically, we probe the residual streams of the LLM. 
Finally, building on the success of prior steering methods \cite{gu-etal-2023-controllable, turner2023activation, marks2024the}, we steer LLM outputs, enabling partial control over whether responses rely on PK or CK.
In summary, our contributions in this paper are as follows:
\begin{itemize}
    \item We expand the investigation of knowledge conflict behavior to the code generation task. To the best of our knowledge, we are the first to investigate knowledge conflict behavior in the context of code generation tasks. 
    \item We construct a framework for testing the LLM's behavior on code generation tasks, and found that the model tends to rely more on PK when the task is easier and the model size is larger.
    \item We propose the use of a probe to detect which type of knowledge the generated answer primarily relies on.
    \item We employ an activation steering method to control whether or not a model uses its parametric knowledge or conflicting knowledge.
\end{itemize}

\section{Related Work}

\paragraph{Context-Memory Conflicts}
Prior work on context-memory conflicts has predominantly focused on single and multi-hop question-answering (QA) tasks \cite{longpreEntityBasedKnowledgeConflicts2021, jin-etal-2024-tug, xie2023adaptive, ying-etal-2024-intuitive, kortukov2024studying, suConflictBankBenchmarkEvaluating2024}. These efforts often focus on crafting more plausible conflicts by ensuring coherent contextual information. However, while this line of research has advanced our understanding of how LLMs handle contradictory knowledge, it remains largely limited to QA tasks. Such focus limits insights into more reasoning-intensive tasks that extend beyond factual retrieval. By examining context-memory conflicts in code generation, we gain a structured yet rich testbed for these broader investigations.

\paragraph{Interpreting Model Behavior}
Gaining insight into why LLMs respond to context-memory conflicts as they do require a closer look at their internal workings. Mechanistic interpretability techniques—such as examining attention heads, probing hidden state activations, and tracing pathways of information flow—have begun to illuminate how different model components store long-term knowledge or incorporate new contextual details. For example, \citet{jin-etal-2024-cutting} demonstrate that selectively disabling certain ``memory heads'' or ``context heads'' can influence a model’s reaction to CK in QA settings. 

Linear probing is a technique used to investigate what type of information is encoded in the representations of a neural model. A linear classifier is trained over the representations to predict specific properties. This allows us to evaluate how relevant the property is in the embeddings or if it is even encoded in them. A linear classifier only performs well if that property is linearly separable in the embedding space, meaning it might be more relevant in that layer \cite{belinkov2022probing}. 

\citealt{tighidet-etal-2024-probing} introduces a linear probing framework that examines LLM activations to determine whether a model relies on PK or CK. The study finds that mid-layer activations, particularly those associated with relational tokens, play a crucial role in determining whether an LLM relies on PK or CK. 
Extending these interpretability methods to the programming domain can uncover both general principles and domain-specific patterns governing a model's behavior in the face of knowledge conflicts.

\section{Framework}\label{sec:framework}
\begin{figure}[ht]
    \centering
    \includegraphics[width=\columnwidth]{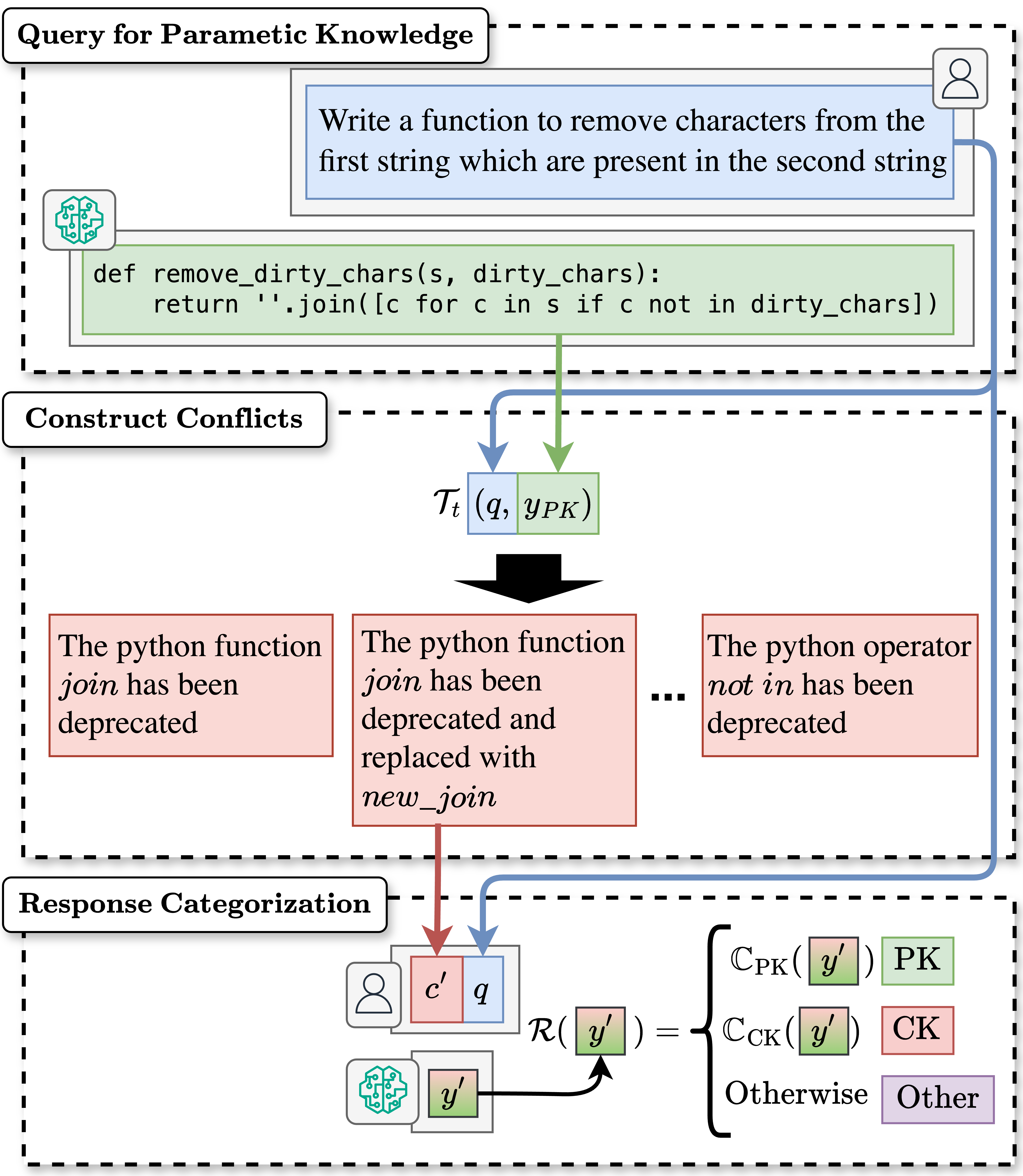}
    \caption{Framework for analyzing context-memory conflicts. \textbf{Query for Parametric Knowledge} elicits the models \(y_{PK}\) for a given query \(q\). \textbf{Construct Conflicts} applies a task-specific template function to create the conflicts from \(q\) and \(y_{PK}\). \textbf{Response Categorization} categorizes the model's response \(y'\) to the combined prompt \((c', q)\)\ either PK, CK, or Other.} 
    \label{fig:framework}
\end{figure}

We begin by establishing a general framework for context-memory conflicts in LLMs, providing a systematic approach to studying how these models handle contradictory information. Figure~\ref{fig:framework} illustrates this framework with an example.

\paragraph{Basic Setup.} Consider an autoregressive language model \(\mathcal{M}\). Given a prompt \( x \), the model produces a token sequence \( y = \mathcal{M}(x) \). We decompose each prompt \( x \) into two disjoint components: a \emph{context} \( c \) and a \emph{query} \( q \), such that \( x = (c, q) \). The context \( c \) provides background information that may influence the model's response, while \( q \) solicits specific information.

\paragraph{Parametric Knowledge.} The model's \emph{parametric knowledge} (PK) is its response to \(q\) in isolation: \(y_{\text{PK}} = \mathcal{M}(q)\). We then denote the model-specific PK for \(q\) as \((q, y_{\text{PK}})\). 

\paragraph{Knowledge Conflicts.} A \emph{knowledge conflict} occurs when a context \( c' \) presents information that contradicts the model's PK for a given query \( q \), denoted as \((c',q)\), where \( c' \) is semantically incompatible with \((q,y_{\text{PK}})\). Note that PK (or CK) does not imply that the response is universally correct; it simply represents the model’s output given the query, without assessing its correctness.

\paragraph{Constructing Conflicts.}
Conflicts are created using a template function:

\[
\mathcal{T}_t: (q, y_{\text{PK}}) \mapsto c',
\]

where \(\mathcal{T}_t\) is a task \(t\) specific template that creates a conflicting context \(c'\) based on \(q\) and \(y_{\text{PK}}\). The $y_\text{PK}$ information is used to generate conflicting information included in the PK.

\paragraph{Response Categorization.} Given a conflicting context prompt \(x' = (c', q)\), let \(y' = \mathcal{M}(x')\) be the model's response. We define a task-specific response classifier \(\mathcal{R}(y')\) that categorizes \(y'\) into three classes:
\[
\mathcal{R}(y') = 
\begin{cases} 
\text{Parametric} & \text{if } \mathbb{C}_{\text{PK}}(y') \\
\text{Conflicting} & \text{if } \mathbb{C}_{\text{CK}}(y') \\
\text{Other} & \text{otherwise}
\end{cases}
\]
where \(\mathbb{C}_{\text{PK}}(y')\) and \(\mathbb{C}_{\text{CK}}(y')\) are task-specific evaluation conditions that determine whether the response aligns with the PK or the CK, respectively. Note that PK (or CK) is not associated with the correctness of a response; it simply serves to categorize the source of the information.

This framework provides a foundation for our systematic investigation of how LLMs handle conflicting information across different tasks and domains.

\section{Datasets and Setup}
\label{sec:datasets}
We extend our framework from Section~\ref{sec:framework} to two different types of tasks: QA and code generation. Full details and examples for both tasks and all datasets are available in Appendix~\ref{sec:appendix:prompt_templates}.
\subsection{QA}
The \emph{World Capitals} dataset is constructed from country-capital pairs, where the query \(q\) asks, ``What is the capital of \texttt{[country]}?'' Because this information is widely known and likely to appear frequently in a model's training data, we also include an \emph{Olympics Winners} dataset, derived from historical Olympic results\footnote{\url{https://github.com/KeithGalli/Olympics-Dataset}}, where the query \(q\) takes the form, ``Who won the gold medal in \texttt{[event description]}?''

For both datasets, we generate conflicting contexts \(c'\) for each query using four types of conflict statements to simulate various scenarios of factual inconsistencies: (1) Default conflicts, (2) Time conflicts, (3) Endorsement conflicts, and (4) Combined Time and Endorsement conflicts. The model's responses under these conflicts are categorized as \textit{Parametric} if they match the no conflict response (\(y_{PK}\)), \textit{Conflicting} if they align with the information in \(c'\), and \textit{Other} otherwise.

\subsection{Code Generation}
For the code generation task, we utilize the EvalPlus \cite{evalplus} dataset, which is comprised of the MBPPPlus (MBPPP) and HumanEvalPlus (HEP) datasets. For this task, the queries \(q\) are the provided prompts that ask the model to generate code to solve a problem in Python.
To create knowledge conflicts for code generation, we first parse the model's parametric code response (\(y_{\text{PK}}\)) to identify the functions and operators used in its response. We construct three types of conflicts \(c'\): Function Deprecation, Operator Deprecation, and Function Replacement. Function and Operator Deprecation conflicts indicate that specific functions or operators used in \(y_{\text{PK}}\) are deprecated. Function Replacement conflicts specify a replacement function for a function used in \(y_{\text{PK}}\).

Responses to conflicting prompts are categorized based on code content analysis. For Function and Operator Deprecation conflicts, a response is categorized as \textit{Conflicting} if it does not contain the deprecated function or operator, and \textit{Parametric} if it does contain the deprecated function or operator. For Function Replacement conflicts, the categorization is more nuanced: a response is \textit{Conflicting} if it contains the replacement function but not the original function; \textit{Parametric} if it contains the original function but not the replacement function; and \textit{Other} if it contains neither or both functions. Responses containing both the original and replacement functions form their own \textit{Other} subcategory. 

\subsection{Environment Setup}
We conducted all experiments using three distinct Llama3 \cite{dubey2024Llama3herdmodels} models following the Llama 3.1 Community Agreement: 1B, 3B, and 8B. The selection of these models allows for a comprehensive evaluation of model performance across different levels of computational and knowledge capabilities.

\section{How are Knowledge Conflicts Treated?}

We discuss the findings on categorized response proportions and attention maps, which provide insights into the model’s behavior when faced with conflicting information.

\subsection{Response Proportion}
\label{sec:response_proportion}

\begin{figure*}[ht]
    \centering

    \begin{subfigure}[b]{0.45\textwidth}
        \includegraphics[width=\linewidth]{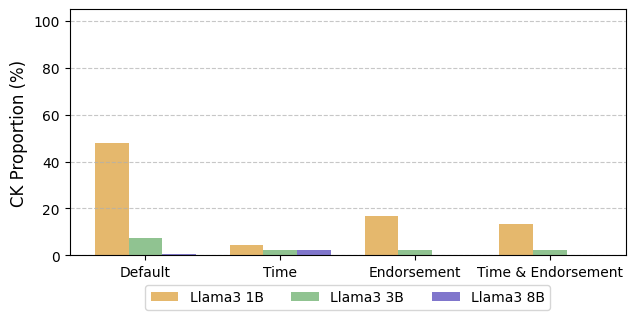}
        \caption{World Capitals Dataset}
        \label{fig:response_proportion:capital}
    \end{subfigure}
    \hfill
    \begin{subfigure}[b]{0.45\textwidth}
        \includegraphics[width=\linewidth]{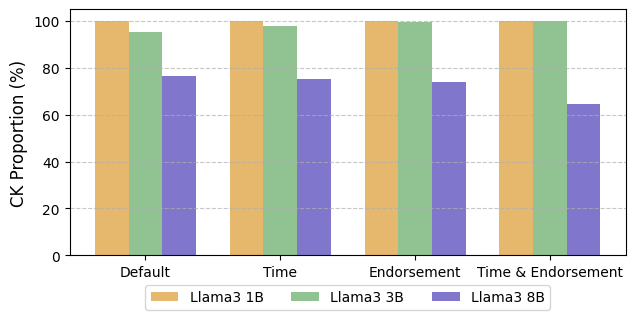}
        \caption{Olympics Winners Dataset}
        \label{fig:response_proportion:olympic}
    \end{subfigure}

    \caption{Proportion of CK-based response of Llama3 model with different model sizes on (a) World Capitals and (b) Olympics Winners dataset. For the detailed results please refer to Appendix \ref{sec:appendix:response_proportion_results}.}
    \label{fig:response_proportion}
\end{figure*}
\label{sec:qa:response_proportion}
\paragraph{QA.} We calculated the percentage of 870 selected responses generated by the LLMs based on PK or CK to observe which type of knowledge the model relies on, depending on different statements, question types, and model sizes. The dataset was constructed by randomly selecting 30 observations and generating 29 conflict pairs for each, resulting in a total of 870 response instances.

The results in Figure~\ref{fig:response_proportion} show the CK proportion patterns vary across Llama3 models' size (1B, 3B, and 8B) and statements on the World Capitals and Olympics Winners datasets, respectively. Notably, when the CK proportion is low, it indicates that the response is generated based on the PK. You can find the detailed percentage of PK and CK-based responses in Appendix~\ref{sec:appendix:response_proportion_results}. In both datasets, as the model size decreases, the proportion of CK-based response is increased for all the statements, which means that when the model size becomes larger, the LLMs tend to rely more on the PK. We believe that this is because with a larger number of parameters, LLMs can remember more information in the model parameters, which strengthens the PK.

When we compare the two datasets, the models generated the responses mostly based on the PK for the World Capitals dataset, while for the Olympics Winners dataset, it is the opposite. This disparity is likely due to the widespread inclusion of World Capitals as a common knowledge base in model training, whereas the Olympics Winners dataset represents more rare information. 

In the context of QA, the proportion of PK, CK, and Others not only reflects the source of knowledge used by the model but also serves as the evaluation metric for accuracy, as correctness is inherently tied to whether the response aligns with PK or CK. 

In summary, when the model size is bigger, and the information is less complicated, the LLMs tend to rely more on the PK. 

\paragraph{Code Generation.}

\begin{figure*}[ht]
    \centering
    \includegraphics[width=0.8\textwidth]{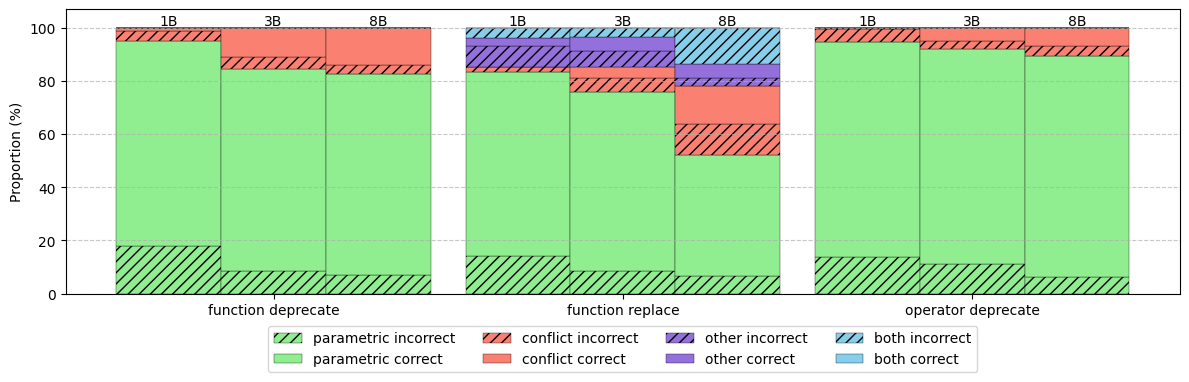}
    \caption{Proportion of code generation responses from the 1B, 3B, and 8B Llama models which are detected as being parametric, conflicting, both, or neither. The responses are separated by perturbation type. Table~\ref{table:code} contains a detailed breakdown}
    \label{fig:code-proportions}
\end{figure*}

We calculated the percentage of code responses generated based on PK or CK and evaluated whether or not the code could be considered correct given that categorization.

Figure~\ref{fig:code-proportions} details the categorization of responses and our correctness evaluation process.  Correctness is assessed by first testing the original code using a test suite.  For \textit{parametric} responses, passing all tests is sufficient for declaring correctness.  For \textit{conflict} responses, we modify the original code according to the conflict type before testing again. Specifically, for deprecation conflicts, we transform deprecated functions or operators into errors. For function replacement conflicts, we additionally transform instances of the new function into the original function. A \textit{conflict} response is deemed correct if the modified code passes all tests.  Responses categorized as \textit{both} and \textit{other} must pass tests under both the original and modified code versions to be considered correct.

We observe that all models answer primarily based on PK across the different perturbation types. The larger models (3B and 8B) generate more \textit{conflict} responses than the 1B model, which indicates that they are more capable of prioritizing CK over PK. The larger models also are more likely to produce correct responses regardless of categorization. 

The operator deprecation task has the lowest rate of CK usage, which is unsurprising considering the low probability that a \texttt{python} operator would ever be deprecated. Interestingly, in the case of function replacement, the 8B model is significantly more likely than the other two models to include both the old function and the new function that is meant to replace it. Additionally, all of the models are less likely to produce \textit{conflict} code that is correct in the case of function replacement as compared with function deprecation. This leads to the interesting conclusion that providing additional CK, even when it would benefit a human programmer by describing an available replacement for a deprecated function, may actually be a detriment to the success of LLMs.
    
In summary, the larger models were more likely to generate \textit{conflict} responses, and are more likely to generate correct code. However, even with the largest model, we find that the models use PK the vast majority of the time, even on these overly simple conflict scenarios.

\subsection{Attention Maps}

Our analysis of self-attention and cross-attention maps reveals distinct patterns in how the models process information. Due to page limitations, the attention map figures are depicted in Appendix~\ref{sec:appendix:attention}. Self-attention maps show that models generally focus on key entities but sometimes shift attention to less relevant words, which does not always align with response trends. This suggests that self-attention may not fully explain how models prioritize information. In contrast, cross-attention maps indicate that models rely more on contextual knowledge in certain cases, while in others, they maintain stronger attention to their internal knowledge. The attention patterns vary depending on the type of input, influencing how models balance external context and stored knowledge. More details can be found in the Appendix~\ref{sec:appendix:attention}.

\section{Can Knowledge Conflicts be Detected?} \label{sec:detection}

To examine whether knowledge conflicts are detectable in the intermediate representations of LLMs, we analyze their internal representations and evaluate how they encode conflicting information.

\paragraph{Method}

Our probing method aims to determine whether knowledge conflicts are embedded within the internal representations of LLMs. To achieve this, we train a simple linear classifier to assess the linear separability of PK and CK. 

Using TransformerLens \cite{nanda2022transformerlens}, we extract the residual stream $h_l$ at layer $l$ for the first generated token from a prompt $p$:
\[
h_l=\text{ResidualStream}_l(p)
\]

A logistic regression classifier serves as the linear probe for each layer, detecting whether the model's response is based on PK or CK:
\[
P(y|h_l)=\sigma(w^Th_l+b)
\]

where $w$ and $b$ are learned parameters, and $\sigma(\cdot)$ represents the sigmoid function.
The predicted label $y$ indicates whether it is PK or CK. We use binary cross-entropy loss as the objective function.

\begin{figure}[tp]
    \centering
    \includegraphics[width=0.47\textwidth]{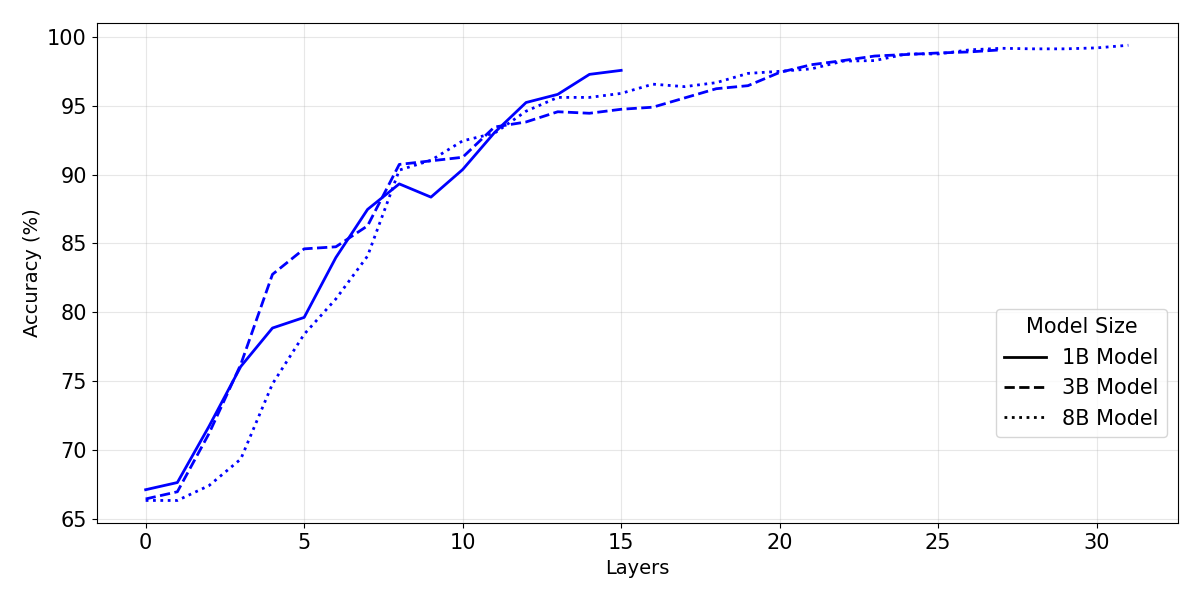}
    \caption{QA Probing Accuracy Across Layers}
    \label{fig:1B_Q/A_linear_probing}
\end{figure}

\begin{figure}[tp]
    \centering
    \includegraphics[width=0.47\textwidth]{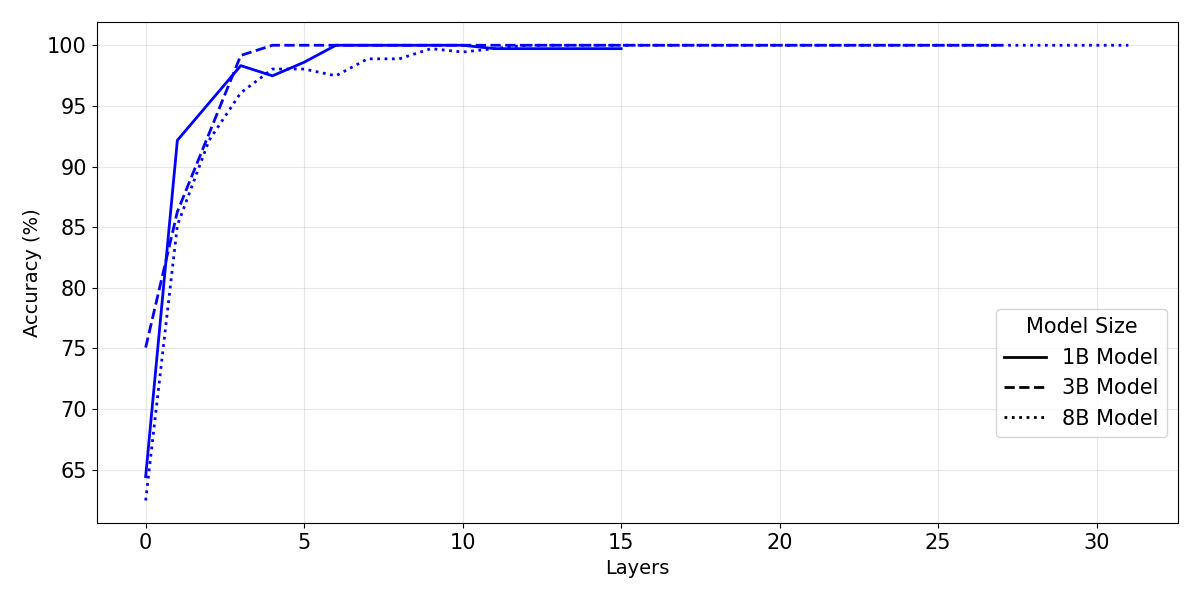}
    \caption{Code Dataset Probing Accuracy Across Layers. Graph created following the structure of Figure \ref{fig:1B_Q/A_linear_probing}.}
    \label{fig:code_linear_probing}
\end{figure}

\paragraph{In-Domain Probe}

We train a linear probe using our two QA datasets, World Capitals and Olympics Winners. Figure \ref{fig:1B_Q/A_linear_probing} illustrates a trend of improvement in accuracy as layers increase. 

We also train a linear probe using the code generation dataset to understand if the LLM can detect knowledge conflicts in other domains. The accuracy results across layers are depicted in Figure \ref{fig:code_linear_probing}, which shows a steeper improvement in accuracy in the earlier layers relative to Figure \ref{fig:1B_Q/A_linear_probing}.

These trends reveal that the models' ability to discriminate between PK and CK is the strongest in later layers,  suggesting that this distinction becomes more pronounced in their deeper embeddings. This observation aligns with \citealp{rogers2021primer} which demonstrated that semantic information is learned and encoded in the deeper layers of BERT \cite{devlin-etal-2019-bert}, and \citealp{marks2024the} which hypothesized a similar phenomenon in auto-regressive LLMs. Assuming this hypothesis holds, we can infer that as semantic information becomes more prominent in the embeddings, the model’s ability to differentiate between PK and CK also improves. This may indicate that the semantic representations in later layers play a crucial role in the models' capacity to detect knowledge conflicts.

\paragraph{Cross-Domain Probe}
While our findings indicate that LLMs can identify knowledge conflicts within a specific domain, this does not necessarily imply that the ability generalizes across domains. To investigate this, we train a linear probe on the QA datasets and test it on the code generation dataset across five different seeds.

\begin{figure}[tp]
    \centering
    \includegraphics[width=0.5\textwidth]{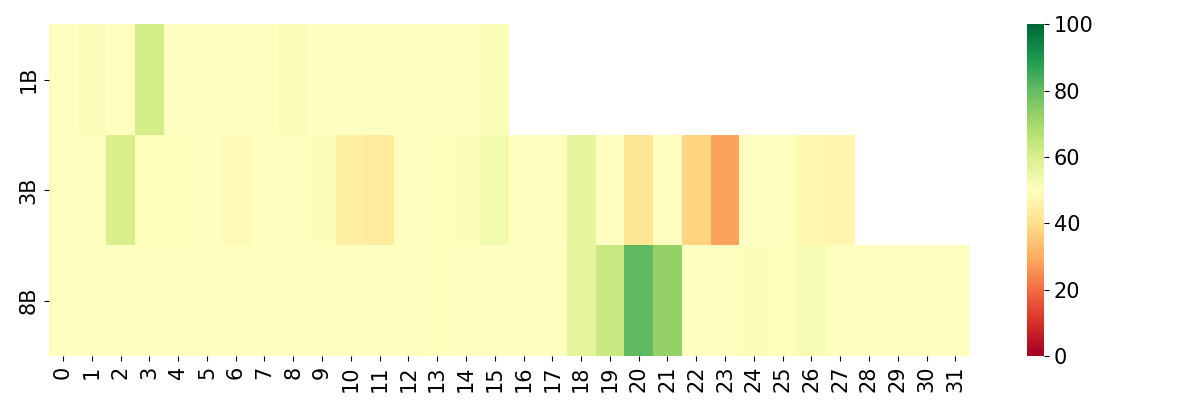}
    \caption{Probing Domain Transfer Accuracy Across Layers.} \label{fig:qa2code_linear_probing}
\end{figure}

Figure \ref{fig:qa2code_linear_probing} shows the results for the average across the five seeds. Accuracy remains close to 50\% overall, with some peaks of statistically significant, non-random classification. This occurs in the early layers of the 1B model, the early and high-mid layers of the 3B model, and the high-mid layers of the 8B model (refer to Appendix \ref{sec:appendix:probe_results}). The strongest distinction ability occurs in the 8B model in layers 18 to 21 with the highest accuracy reaching 80.65\%.
These results suggest that the ability to distinguish between PK and CK does transfer across domains, but is most pronounced in larger models.

\begin{figure}[tp]
    \centering
    \includegraphics[width=0.5\textwidth]{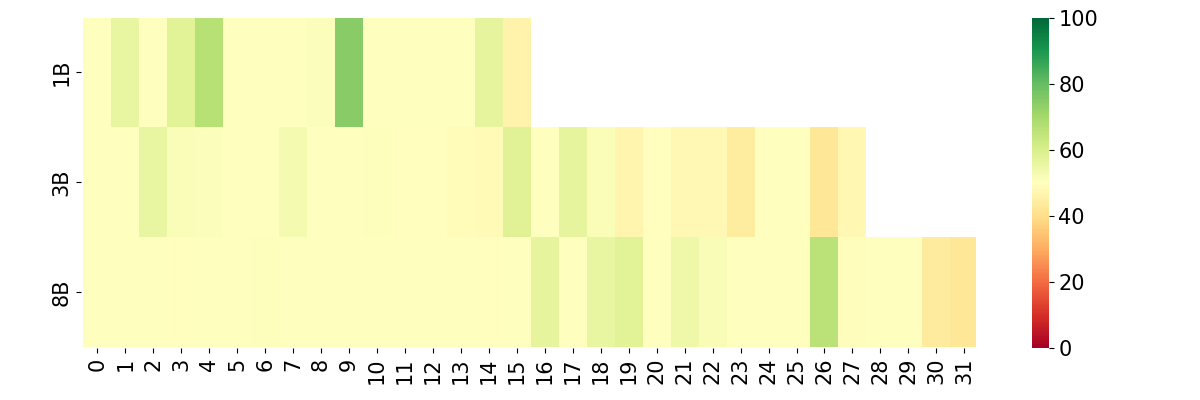}
    \caption{Nonlinear Probing Domain Transfer Accuracy Across Layers.}
    \label{fig:qa2code_nonlinear_probing}
\end{figure}

If knowledge conflicts are more complexly embedded, a non-linear probe may be better suited for detection. To investigate this, we train a non-linear probe using a two-layer MLP with ReLU activation. The results of this probe, shown in Figure \ref{fig:qa2code_nonlinear_probing}, demonstrate a similar trend of hovering around 50\% with some statistically significant peaks. However, unlike the linear probe, the 1B model shows notable peaks, with the highest occurring at layer nine. 

Statistically significant peaks are observed in the early and high-mid layers of the 1B and 3B models, as well as in the mid-high layers of the 8B model. Notably, the non-linear probe may have overfit to the linearly separable aspects of knowledge conflict detection, reducing its effectiveness in the 8B model.

Since the probes perform very well in-domain across all model sizes but tend to struggle to generalize well across domains, domain-specific factors must influence how this distinction is encoded. This suggests that a general concept of knowledge conflict is not strongly linearly embedded in auto-regressive LLMs. Instead, the model relies primarily on domain-specific and semantic factors to distinguish PK from CK, rather than encoding knowledge conflicts as a general concept. However, as model size increases, evidence of a more generalizable knowledge conflict representation begins to emerge.

\section{Steering Responses Under Conflicts}
\label{sec:steering}
Following the results from Section \ref{sec:detection}, we examine the possibility of \textit{steering} a model to give responses that align with either its PK or CK from the context when the model is faced with a context-memory conflict.

\paragraph{Method}
Our steering method aims to influence the model's output by modifying its internal activations. To do this, we construct a steering vector $\mathbf{s}$ which we either add or subtract from the model's activations to bias its behavior.

As in Section \ref{sec:detection}, we focus solely on this residual stream and choose the layer for which the probe had the highest test accuracy.

Following \citet{marks2024the}, we construct our steering vector using the difference in mean activations between conflicting prompts ($X'$) and regular prompts ($X$):
\[
\mathbf{u} = \frac{1}{|X'|} \sum_{x' \in X'} \mathbf{a}(x'), \quad
\mathbf{v} = \frac{1}{|X|} \sum_{x \in X} \mathbf{a}(x)
\]
where $\mathbf{a}(x)$ represents the residual stream activation for prompt $x$. The activation difference $\mathbf{u} - \mathbf{v}$ represents the typical change in activations when the model encounters conflicting information. We project this difference onto the probe's normalized weight vector to obtain the steering vector $\mathbf{s}$:
\[
\mathbf{s} = \left( (\mathbf{u} - \mathbf{v}) \cdot \boldsymbol{\theta} \right) \frac{\boldsymbol{\theta}}{\|\boldsymbol{\theta}\|_2}
\]
During inference, we apply $\mathbf{s}$ to the model's residual stream activations:
\[
\mathbf{a}'(x) = \mathbf{a}(x) \pm \mathbf{s}
\]

\paragraph{Metrics}
To evaluate steering effectiveness, we measure how the rate at which we successfully change which knowledge source the model uses. For a set of conflicting prompts $X'$, let $X'_{CK} = \{x' \in X' | \mathbb{C}_{\text{CK}}(\mathcal{M}(x')) \}$ be the set of prompts where the model's response aligns with the conflicting context. Similarly, let $X'_{PK} = \{x' \in X' | \mathbb{C}_{\text{PK}}(\mathcal{M}(x')) \}$ be the set of prompts where the model's response aligns with parametric knowledge. We then define the success rate towards PK ($S_{PK}$) and the success rate towards CK ($S_{CK}$) as:
\[
S_{PK} = \frac{|\{x' \in X' \setminus X'_{PK} \mid \mathbb{C}_{\text{PK}}(\mathcal{M}^{+\mathbf{s}}(x'))\}|}{|X' \setminus X'_{PK}|}
\]
\[
S_{CK} = \frac{|\{x' \in X' \setminus X'_{CK} \mid \mathbb{C}_{\text{C}}(\mathcal{M}^{-\mathbf{s}}(x'))\}|}{|X' \setminus X'_{CK}|}
\]
Here, $\mathcal{M}^{+\mathbf{s}}(x')$ and $\mathcal{M}^{-\mathbf{s}}(x')$ denote the model's response to prompt $x'$ after applying the positive and negative steering vector, respectively. $\mathbb{C}_{\text{CK}}$ and \(\mathbb{C}_{\text{PK}}\) are the evaluation conditions for conflicting and parametric responses as described in Section \ref{sec:framework}.

\begin{table}[htbp]
{\small
  \centering
  \caption{Steering Success Rates}
  \label{tab:steeering_results}
  \scalebox{0.9}{
    \begin{tabular}{
      >{\centering\arraybackslash}m{1cm} 
      >{\centering\arraybackslash}m{1cm} 
      >{\centering\arraybackslash}m{1.3cm} 
      >{\centering\arraybackslash}m{1.3cm} 
      >{\centering\arraybackslash}m{1.3cm} 
    }
      \toprule
       \textbf{Task} & \textbf{Model} & \textbf{\(S_{PK}\) (\(\uparrow\))} & \textbf{\(S_{CK}\) (\(\uparrow\))} & \textbf{\(S_{avg}\) (\(\uparrow\))} \\
      \midrule
      \multirow{3}{*}{WC}  & 1B  & 0.064 & 0.024 & 0.039 \\
                           & 3B  & 0.214 & 0.237 & 0.231 \\
                           & 8B  & 0.200 & 0.010 & 0.020 \\
      \midrule
      \multirow{3}{*}{OW}  & 1B & 0.000 & -- & 0.000 \\
                           & 3B & 0.010 & 0.200 & 0.019 \\
                           & 8B & 0.038 & 0.240 & 0.117 \\
      \midrule
      \multirow{3}{*}{HEP} & 1B & 0.313 & 0.163 & 0.185 \\
                           & 3B & 0.318 & 0.171 & 0.202 \\
                           & 8B & 0.261 & 0.105 & 0.138 \\
      \midrule
      \multirow{3}{*}{MBPPP} & 1B & 0.000 & 0.239 & 0.202 \\
                            & 3B & 0.412 & 0.060 & 0.119 \\
                            & 8B & 0.200 & 0.057 & 0.084 \\
      \midrule
      \multirow{3}{*}{\shortstack{QA\\to\\Code}} & 1B & 0.044 & 0.023 & 0.028 \\
                            & 3B & 0.100 & 0.012 & 0.029 \\
                            & 8B & 0.250 & 0.092 & 0.126 \\
      \bottomrule
    \end{tabular}
  }}
  \par\smallskip
    \small Values missing where either \(X'_{CK}\) or \(X'_{PK}\) are empty. \(S_{avg}\) is the overall steering success rate.
\end{table}

\paragraph{Analysis}
Table \ref{tab:steeering_results} reveals that activation-based steering achieves varying degrees of success, highlighting a nuanced relationship between task and model characteristics. While the observed steering success rates are not uniformly high, they highlight the potential for targeted interventions in model behavior at the activation level, even generalizing across domains as seen in the 8B model, achieving an overall steering success rate (\(S_{avg}\)) of 0.126 when probes trained on QA were used to steer conflicts in the Code task. Notably, the effectiveness of steering is significantly influenced by both the knowledge source we aim to steer towards and the specific task.

Steering towards PK (\(S_{PK}\)) demonstrates comparatively higher success rates across several tasks, particularly for WC, HEP, MBPPP (excluding the 1B model), and QA to Code Generation. Conversely, steering towards CK (\(S_{CK}\)) appears more effective for the OW task. We hypothesize that this arises from the differences in the strength knowledge representations formed during pre-training. Tasks like WC, HEP, and QA to Code likely involve information (e.g., world capitals and Python code) that is prevalent in typical pre-training corpora. The high exposure to this information leads to the formation of robust representations within the model's PK that are more amenable to steering toward this dominant knowledge source. In contrast, the OW task, which queries for knowledge less likely to be encountered during pre-training, might engage less confidently held parametric representations. These weaker representations result in steering toward CK being easier than steering toward PK. For OW, the higher \(S_{CK}\) values suggest that in the absence of strong parametric priors, the model is more readily biased toward the context.

Table \ref{tab:steeering_results} also reveals model-task-specific scaling trends in overall steering success. For MBPPP, we observe a decrease in steering efficacy with increasing model size, while OW exhibits the opposite trend. We propose that for tasks like MBPPP, where the relevant information is likely well-represented and confidently encoded in larger models, the increased model capacity primarily serves to solidify the model's PK. This solidity makes these larger models more inert to activation-based steering, as their responses are already strongly biased by their learned weights. Conversely, for tasks like OW, where the information is less prevalent in pre-training, the enhanced instruction-following capabilities in the larger models can be leveraged in steering to more effectively guide responses, particularly towards the CK, which is less in conflict with their weaker parametric priors for this task.


\section{Conclusions and Future Work}

In this paper, we analyze how LLMs handle context-memory conflicts across QA and Code Generation tasks. Through systematic analysis of model responses across various conflicting scenarios, we uncover how models attend to contradictory information and develop strategies to reliably detect and control knowledge conflicts.

Our analysis shows that models' adherence to parametric knowledge strongly correlates with their confidence in stored information -- demonstrated by high resistance to conflicts in well-known domains (World Capitals) versus greater flexibility with less certain knowledge (Olympics Winners). Additionally, we show that as the model size increases, the models align more closely to the parametric knowledge. 
Probing techniques demonstrate that within the parameters of a large enough model exists the general concept of knowledge conflicts. Steering effectiveness is highly dependent on model size, task domain, and steering direction.

Future work should explore additional domains beyond QA and code generation, develop better predictive methods for when models will favor PK versus conflicting knowledge, and investigate how model architecture affects conflict resolution strategies. These insights will be crucial for developing more reliable AI systems that can effectively navigate knowledge conflicts while maintaining high performance across diverse tasks. 
Additionally, we plan to explore incorporating memory via multi-turn conversations into the analysis, allowing for a more comprehensive understanding of response adaptability and enabling users to steer LLM outputs more effectively.

\section{Limitations}
Several limitations in our study need to be addressed to enhance its real-world applicability. First, our probing analysis is limited in identifying which layers respond to CK. However, we have yet to compare advanced techniques, such as causal probing \cite{meng2022locating} and attention flow analysis \cite{kovaleva-etal-2019-revealing}, to better understand how LLMs handle knowledge conflicts in QA and code generation tasks. While we extend our analysis from simple QA to code generation, further expanding task diversity, incorporating more nuanced knowledge conflicts, and controlling for confounding factors in dataset design could improve the robustness of our findings.

\section{Ethical Considerations}
This work does not pose any ethical issues. All code and datasets used in this work are publicly available and utilized in compliance with their respective licenses. We also declare that all authors of this paper acknowledge the ACM Code of Ethics and honor the code of conduct.

\section{Acknowledgements}

This work used the Delta system at the National Center for Supercomputing Applications through allocation CIS240772 from the Advanced Cyberinfrastructure Coordination Ecosystem: Services \& Support (ACCESS) program, which is supported by National Science Foundation grants \#2138259, \#2138286, \#2138307, \#2137603, and \#2138296 \cite{boerner2023a}.

\bibliography{main}

\clearpage
\appendix

\section{Prompt Templates}\label{sec:appendix:prompt_templates}
Detailed templates for the World Capitals and Olympics Winners datasets are provided here. The template we used for the World Capitals Dataset is as follows:

\begin{enumerate}
\item \textbf{Default Statement}: A conflicting statement of the form, \textit{“The capital of [country] is [capital].”}
\item \textbf{Time-sensitive Statement}: A temporally framed assertion, such as \textit{“[Capital] has [temporal verb phrase] become the capital of [country].”} The temporal verb phrases used include \textit{“is now”}, \textit{“will now be”}, \textit{“has become”}, \textit{“has just become”}, and \textit{“has just now become”} which were randomly selected to introduce time-based discrepancies.
\item \textbf{Endorsement Statement}: A statement involving a figure of authority, e.g., \textit{“The president of [country] has announced that [capital] is the capital of [country].”}
\item \textbf{Combined Time and Endorsement Statement}: A compound statement combining temporal and authoritative elements, such as \textit{“The president of [country] has announced that [capital] [temporal verb phrase] the capital of [country].”}
\item \textbf{Query}: The user prompt to feed into the LLM after introducing the conflict. \textit{“What is the [capital] of [country]?”}
\end{enumerate}

The Olympics Winners dataset contains more complicated information. This includes [person], which refers to the name of the athlete; [discipline], representing the specific sport; [category], indicating the classification within the sport; [event], specifying the particular competition; [year], denoting the year in which the event took place; and [season], distinguishing between the Summer and Winter Olympics. A detailed template we used for the Olympics Winners Dataset is as follows:
\begin{enumerate}
\item \textbf{Default Statement}: A conflict statement of the form, \textit{“[Person] is the gold medal winner in the [discipline] [category] [event] event at the [year] [season] Olympics.”}
\item \textbf{Time-sensitive Statement}: A temporally framed assertion, such as \textit{”[Person] [temporal verb phrase] the new gold medal winner in the [discipline] [category] [event] event at the [year] [season] Olympics as of [temporal expression].”} The same temporal verb phrases are used as World Capitals dataest. Additionally, we introduced temporal expressions such as \textit{“just”}, \textit{“yesterday”}, \textit{“today”}, and \textit{“last year”} to provide more precise time-related information.
\item \textbf{Endorsement Statement}: A statement involving a figure of authority, e.g., \textit{“The president of the IOC has announced that due to a doping scandal [person] is the gold medal winner in the [discipline] [category] [event] event at the [year] [season] Olympics.”}
\item \textbf{Combined Time and Endorsement Statement}: A compound statement combining temporal and authoritative elements, such as \textit{“The president of the IOC has announced that due to a doping scandal [person] [temporal verb phrase] the new gold medal winner in the [discipline] [category] [event] event at the [year] [season] Olympics as of [temporal phrase].”}
\item \textbf{Query}: The user prompt to feed into the LLM after introducing the conflict. \textit{“Who is the gold medal winner in the [discipline] [category] [event] event at the [year] [season] Olympics”}
\end{enumerate}

The templates used for code generation contain both a randomly selected introductory statement as well as a conflict statement of a given type (e.g. function replacement). Here are the introductory statements:

\begin{enumerate}
    \item \textbf{Default Statement}: A conflict statement of the form \textit{"The Python [conflict Statement]"}
    \item \textbf{Imagination statement}: A thought-experiment inspired statement, such as \textit{"You are working in a language that is like Python, except the [conflict statement]"}
    \item \textbf{Update Statement}: A statement framed under the very real possibility of breaking changes in updates. \textit{"In the most recent version of Python, the [conflict statement]"}
\end{enumerate}

Here are the conflict statements for the different types of conflict:

\begin{enumerate}
    \item \textbf{Function Deprecate}: \textit{"[introductory statement] function [function] has been deprecated and removed, meaning it can no longer be called."}
    \item \textbf{Function Replacement}: \textit{"[introductory statement] function [function] has been deprecated, but has been replaced with new\_[function] which has an identical implementation and signature to [function]"}
    \item \textbf{Operator Deprecate}: \textit{"[introductory statement] operator [operator] has been deprecated and removed, meaning it can no longer be used."}
\end{enumerate}

\section{Experimental Design}
\label{sec:appendix:exp_design}

\subsection{Randomness}
Data splits and models were generated using seeds: 1, 10, 42, 99, 777.

\subsection{Datasets}
\begin{itemize}
    \item World Capitals: 1108 statements (871 conflict, 237 true)
    \item Olympics Winners: 2267 statements (1398 conflict, 871 true)
    \item Code: 468 statements (234 conflict, 234 true)
\end{itemize}

\subsection{Probing}
Probing used an 80/20 train/test split.

\subsection{Steering}
Steering used the best-performing probe (by test accuracy) for each model and dataset.  100 model-specific conflict prompts were used for each steering experiment.

\section{Detailed Results of Response Proportion}
\label{sec:appendix:response_proportion_results}
A detailed number of response proportion results of QA tasks of Figure~\ref{fig:response_proportion} are presented in Table~\ref{table:capital} and Table~\ref{table:olympic}. The code generation results in Figure~\ref{fig:code-proportions} are presented in Table~\ref{table:olympic}.

\begin{table*}[htbp]  
\centering
\setlength{\tabcolsep}{10pt}  
{
\resizebox{\textwidth}{!}{
\begin{tabular}{l rrr rrr rrr}
    \toprule
        \multirow{2}{*}{\textbf{Statement}} & \multicolumn{3}{c}{\textbf{Llama3 8B}} & \multicolumn{3}{c}{\textbf{Llama3 3B}} & \multicolumn{3}{c}{\textbf{Llama3 1B}}  \\ 
        \cmidrule(lr){2-4} \cmidrule(lr){5-7} \cmidrule(lr){8-10}
        ~ & \multicolumn{1}{c}{\textbf{CK}} & \multicolumn{1}{c}{\textbf{PK}} & \multicolumn{1}{c}{\textbf{Others}} & \multicolumn{1}{c}{\textbf{CK}} & \multicolumn{1}{c}{\textbf{PK}} & \multicolumn{1}{c}{\textbf{Others}} & \multicolumn{1}{c}{\textbf{CK}} & \multicolumn{1}{c}{\textbf{PK}} & \multicolumn{1}{c}{\textbf{Others}}  \\ 
        \midrule
        Default & 
        0.46 & \textbf{99.08} & 0.46 & \textbf{7.59} & 88.97 & 3.45 & \textbf{47.70} & 35.29 & 17.01 \\
        Time & \textbf{2.41} & 95.86 & 1.72 & 2.30 & 90.46 & \textbf{7.24} & 4.25 & \textbf{65.63} & 30.11 \\
        Endorsement & 0.00 & 97.59 & \textbf{2.41} & 2.30 & \textbf{93.45} & 4.25 & 16.78 & 52.53 & \textbf{30.69} \\
        Time \& Endorsement & 0.11 & 98.16 & 1.72 & 2.30 & 90.46 & \textbf{7.24} & 13.56 & 58.97 & 27.47 \\ \bottomrule

\end{tabular}}}
\caption{Categorized response ratio of Llama3 model with different model sizes on World capital dataset. C, P, and O indicate conflicting, parametric, and other responses. All numbers are in \% scale.}
\label{table:capital}
\end{table*}

\begin{table*}[htbp]  
\centering
\setlength{\tabcolsep}{10pt}  
{
\resizebox{\textwidth}{!}{
\begin{tabular}{ l rrr rrr rrr}
    \toprule
        \multirow{2}{*}{\textbf{Statement}} & \multicolumn{3}{c}{\textbf{Llama3 8B}} & \multicolumn{3}{c}{\textbf{Llama3 3B}} & \multicolumn{3}{c}{\textbf{Llama3 1B}}  \\ 
        \cmidrule(lr){2-4} \cmidrule(lr){5-7} \cmidrule(lr){8-10}
        ~ & \multicolumn{1}{c}{\textbf{CK}} & \multicolumn{1}{c}{\textbf{PK}} & \multicolumn{1}{c}{\textbf{Others}} & \multicolumn{1}{c}{\textbf{CK}} & \multicolumn{1}{c}{\textbf{PK}} & \multicolumn{1}{c}{\textbf{Others}} & \multicolumn{1}{c}{\textbf{CK}} & \multicolumn{1}{c}{\textbf{PK}} & \multicolumn{1}{c}{\textbf{Others}}  \\ 
        \midrule
        Default & 76.67 & 9.31 & 14.02 & 95.17 & 0.34 & 4.48 & 100.00 & 0.00 & 0.00 \\ 
        Time & 75.06 & 7.47 & 17.47 & 97.93 & 0.46 & 1.61 & 100.00 & 0.00 & 0.00  \\
        Endorsement & 73.79 & 8.39 & 17.82 & 99.54 & 0.00 & 0.46 & 100.00 & 0.00 & 0.00 \\
        Time \& Endorsement & 64.6 & 9.31 & 26.09 & 100.00 & 0.00 & 0.00 & 100.00 & 0.00 & 0.00 \\
        \bottomrule
\end{tabular}}}
\caption{Categorized response ratio of Llama3 model with different model sizes on Olympics Winners dataset. C, P, and O indicate conflicting, parametric, and other responses. All numbers are in \% scale.}
\label{table:olympic}
\end{table*}

\begin{table*}[htbp]  
\centering
{\normalsize
\begin{tabular}{ccrrrrrrrr}
\toprule
\multicolumn{1}{c}{\multirow{2}{*}{\textbf{Model}}} &
\multicolumn{1}{c}{\multirow{2}{*}{\textbf{Perturbation Kind}}} 
& \multicolumn{2}{c}{\textbf{Both}} & \multicolumn{2}{c}{\textbf{Parametric}} & \multicolumn{2}{c}{\textbf{Conflict}} & \multicolumn{2}{c}{\textbf{Other}} \\
\cmidrule(lr){3-4} \cmidrule(lr){5-6} \cmidrule(lr){7-8} \cmidrule(lr){9-10}
~ & ~ & \multicolumn{1}{c}{\textbf{\xmark}} & \multicolumn{1}{c}{\textbf{\cmark}} & \multicolumn{1}{c}{\textbf{\xmark}} & \multicolumn{1}{c}{\textbf{\cmark}} & \multicolumn{1}{c}{\textbf{\xmark}} & \multicolumn{1}{c}{\textbf{\cmark}} & \multicolumn{1}{c}{\textbf{\xmark}} & \multicolumn{1}{c}{\textbf{\cmark}}
\\
 \midrule
\multirow[c]{3}{*}{1B} & function deprecate & 0.00 & 0.00 & 17.95 & 76.92 & 3.85 & 1.28 & 0.00 & 0.00 \\
 & function replace & 3.85 & 0.00 & 14.10 & 69.23 & 1.92 & 0.00 & 7.69 & 3.21 \\
 & operator deprecate & 0.00 & 0.00 & 13.82 & 80.92 & 4.61 & 0.66 & 0.00 & 0.00 \\
 \midrule
\multirow[c]{3}{*}{3B} & function deprecate & 0.00 & 0.00 & 8.61 & 75.82 & 4.51 & 11.07 & 0.00 & 0.00 \\
 & function replace & 3.69 & 0.00 & 8.61 & 67.21 & 5.33 & 4.10 & 6.15 & 4.92 \\
 & operator deprecate & 0.00 & 0.00 & 11.26 & 80.63 & 3.15 & 4.95 & 0.00 & 0.00 \\
\midrule
\multirow[c]{3}{*}{8B} & function deprecate & 0.00 & 0.00 & 6.88 & 75.72 & 3.26 & 14.13 & 0.00 & 0.00 \\
 & function replace & 13.77 & 0.00 & 6.52 & 45.65 & 11.59 & 14.13 & 3.26 & 5.07 \\
 & operator deprecate & 0.00 & 0.00 & 6.10 & 83.33 & 3.66 & 6.91 & 0.00 & 0.00 \\
\bottomrule
\end{tabular}
\caption{Categorized response ratio of Llama3 model with different model sizes on our evalplus-based code generation task with different kinds of perturbations. All numbers are in \% scale. \cmark and \xmark~indicate the correct and incorrect responses, respectively.}
\label{table:code}
}
\end{table*}

\section{Detailed Results of Probes}
\label{sec:appendix:probe_results}

This section presents detailed results for all probing tasks, including QA, Code, Linear Domain Transfer (LDT), and Nonlinear Domain Transfer (NLDT). Results for the 1B, 3B, and 8B models are shown in Tables \ref{tab:1b_probe_results}, \ref{tab:3b_probing_results}, and \ref{tab:8b_probing_results}, respectively. A t-test has been applied to both the LDT and NLDT tasks.
\begin{table}[h]
{\small
\begin{tabular}{c rrrr}
\toprule
\multirow{2}{*}{\textbf{Layer}} & \multicolumn{4}{c}{\textbf{Llama 1B Task Accuracy}}                                           \\ 
\cmidrule(lr){2-5}
~                       & \multicolumn{1}{c}{\textbf{QA}}    & \multicolumn{1}{c}{\textbf{Code}}  & \multicolumn{1}{c}{\textbf{LDT}} & \multicolumn{1}{c}{\textbf{NLDT}} \\ 
\midrule

0                                           & 67.15 & 64.43 & 50.00±0.00                    & 50.00±0.00                     \\ 
1                                           & 67.67 & 92.16 & 50.95±1.49             & 55.82±1.49                \\
2                                           & 71.75 & 95.24 & 50.00±0.00                   & 50.00±0.00                      \\
3                                           & 76.08 & 98.32 & \textbf{60.82±1.29}    & 57.68±1.29                \\
4                                           & 78.87 & 97.48 & 50.00±0.50                 & \textbf{66.78±0.50}        \\
5                                           & 79.64 & 98.60  & 50.00±0.00                   & 50.00±0.00                      \\
6                                           & 83.98 & 100.00   & 50.00±0.11                & 50.19±0.11                \\
7                                           & 87.47 & 100.00   & 50.00±0.00                   & 50.00±0.00                      \\
8                                           & 89.32 & 100.00   & 51.44±0.13             & 50.86±0.13                \\
9                                           & 88.36 & 100.00   & 49.95±1.90              & \textbf{74.24±1.90}        \\
10                                          & 90.36 & 100.00   & 50.00±0.00                   & 50.00±0.00                      \\
11                                          & 93.03 & 99.72 & 50.00±0.00                   & 50.00±0.00                      \\
12                                          & 95.22 & 99.72 & 50.00±0.00                   & 50.00±0.00                      \\
13                                          & 95.81 & 99.72 & 50.00±0.00                   & 50.00±0.00                      \\
14                                          & 97.26 & 99.72 & 50.24±0.49             & 56.44±0.49                \\
15                                          & 97.55 & 99.72 & 51.53±0.52             & 46.4±0.52                 \\
\bottomrule
\end{tabular}}
\caption{Probing results for LLaMA-1B. Bolded values indicate statistical significance (p \textless 0.01). LDT and NLDT denote linear domain transfer and nonlinear domain transfer, respectively.}
\label{tab:1b_probe_results}
\end{table}

\begin{table}[h]
{\small
\begin{tabular}{c rrrr}
\toprule
\multirow{2}{*}{\textbf{Layer}} & \multicolumn{4}{c}{\textbf{Llama 3B Task Accuracy}}                \\ 
\cmidrule(lr){2-5}
                       & \multicolumn{1}{c}{\textbf{QA}}    & \multicolumn{1}{c}{\textbf{Code}}  & \multicolumn{1}{c}{\textbf{LDT}}    & \multicolumn{1}{c}{\textbf{NLDT}}  \\ 
                       \midrule
0                      & 66.48 & 75.07 & 50.00±0.00                & 50.00±0.00                \\
1                      & 67.00    & 86.27 & 50.34±0.00             & 50.00±0.00                \\
2                      & 71.23 & 92.72 & \textbf{60.15±0.97} & \textbf{55.63±0.97} \\
3                      & 76.23 & 99.16 & 50.45±0.33          & 51.20±0.33           \\
4                      & 82.76 & 100.00   & 49.29±0.12          & 51.04±0.12          \\
5                      & 84.61 & 100.00   & 50.10±0.00              & 50.00±0.00                \\
6                      & 84.76 & 100.00   & 48.27±0.17          & 50.29±0.17          \\
7                      & 86.28 & 100.00   & 50.00±0.70              & \textbf{52.96±0.70}  \\
8                      & 90.73 & 100.00   & 50.00±0.00                & 50.00±0.00                \\
9                      & 90.99 & 100.00   & 50.83±0.00             & 50.00±0.00                \\
10                     & 91.25 & 100.00   & 44.78±0.40           & 50.68±0.40           \\
11                     & 93.44 & 100.00   & 43.55±0.04          & 49.99±0.04          \\
12                     & 93.81 & 100.00   & 50.22±0.01          & 49.99±0.01          \\
13                     & 94.55 & 100.00   & 49.51±0.18          & 49.01±0.18          \\
14                     & 94.44 & 100.00   & 50.94±0.54          & 48.35±0.54          \\
15                     & 94.73 & 100.00   & \textbf{53.41±0.60}  & \textbf{58.03±0.60}  \\
16                     & 94.88 & 100.00   & 49.71±0.09          & 50.21±0.09          \\
17                     & 95.55 & 100.00   & 50.35±0.62          & \textbf{56.25±0.62} \\
18                     & 96.22 & 100.00   & \textbf{56.15±0.73} & 51.25±0.73          \\
19                     & 96.44 & 100.00   & 49.80±0.22           & 46.57±0.22          \\
20                     & 97.40  & 100.00   & 42.03±0.13          & 49.84±0.13          \\
21                     & 97.96 & 100.00   & 49.94±0.37          & 47.73±0.37          \\
22                     & 98.26 & 100.00   & 37.66±0.31          & 48.03±0.31          \\
23                     & 98.59 & 100.00   & 28.41±0.74          & 43.87±0.74          \\
24                     & 98.70  & 100.00   & 49.99±0.01          & 50.05±0.01          \\
25                     & 98.81 & 100.00   & 50.00±0.00                & 50.00±0.00                \\
26                     & 98.89 & 100.00   & 47.46±1.40           & 42.49±1.40           \\
27                     & 99.04 & 100.00   & 46.78±0.38          & 47.37±0.38  \\ \bottomrule 
\end{tabular}}
\caption{Probing results for LLaMA-3B. Bolded values indicate statistical significance (p \textless 0.01).}
\label{tab:3b_probing_results}
\end{table}

\newpage
\begin{table}[h]
{\small
\resizebox{\linewidth}{!}{
\begin{tabular}{c rrrr}
\toprule
\multirow{2}{*}{\textbf{Layer}} & \multicolumn{4}{c}{\textbf{Llama 8B Task Accuracy}} \\
\cmidrule(rl){2-5}
                       & \multicolumn{1}{c}{\textbf{QA}}    & \multicolumn{1}{c}{\textbf{Code}}  & \multicolumn{1}{c}{\textbf{LDT}}     & \multicolumn{1}{c}{\textbf{NLDT}}  \\ \midrule
0                      & 66.37 & 62.46 & 50.00±0.00                & 50.00±0.00                \\
1                      & 66.37 & 85.15 & 50.00±0.00                & 50.00±0.00                \\
2                      & 67.45 & 92.16 & 50.00±0.16             & 50.03±0.16          \\
3                      & 69.37 & 96.08 & 50.00±0.18             & 49.69±0.18          \\
4                      & 74.82 & 98.04 & 50.00±0.00                & 50.00±0.00                \\
5                      & 78.42 & 98.04 & 50.00±0.00                & 50.00±0.00                \\
6                      & 80.98 & 97.48 & 49.75±0.06          & 50.40±0.06           \\
7                      & 84.09 & 98.88 & 50.00±0.00                & 50.00±0.00                \\
8                      & 90.32 & 98.88 & 50.00±0.00                & 50.00±0.00                \\
9                      & 91.03 & 99.72 & 50.00±0.00                & 50.00±0.00                \\
10                     & 92.44 & 99.44 & 50.00±0.00                & 50.00±0.00                \\
11                     & 92.99 & 99.72 & 50.00±0.00                & 50.00±0.00                \\
12                     & 94.59 & 100.00   & 50.00±0.00                & 50.00±0.00                \\
13                     & 95.59 & 100.00   & 49.27±0.00             & 50.00±0.00                \\
14                     & 95.59 & 100.00   & 50.02±0.09          & 49.88±0.09          \\
15                     & 95.88 & 100.00   & 50.00±0.00                & 50.00±0.00                \\
16                     & 96.55 & 100.00   & 50.00±0.55             & \textbf{56.42±0.55} \\
17                     & 96.37 & 100.00   & 50.00±0.00                & 50.00±0.00                \\
18                     & 96.66 & 100.00   & \textbf{56.48±0.79} & \textbf{55.54±0.79} \\
19                     & 97.33 & 100.00   & \textbf{63.07±1.58} & \textbf{57.78±1.58} \\
20                     & 97.48 & 100.00   & \textbf{80.65±0.04} & 50.04±0.04          \\
21                     & 97.66 & 100.00   & \textbf{72.66±0.79} & \textbf{54.38±0.79} \\
22                     & 98.22 & 100.00   & 50.00±0.64             & 51.69±0.64          \\
23                     & 98.26 & 100.00   & 50.00±0.06             & 50.10±0.06           \\
24                     & 98.74 & 100.00   & 51.14±0.06          & 50.13±0.06          \\
25                     & 98.74 & 100.00   & 50.67±0.02          & 50.24±0.02          \\
26                     & 99.04 & 100.00   & 52.10±0.97           & \textbf{65.70±0.97}  \\
27                     & 99.15 & 100.00   & 49.64±0.32          & 49.31±0.32          \\
28                     & 99.11 & 100.00   & 50.00±0.16             & 50.25±0.16          \\
29                     & 99.11 & 100.00   & 50.00±0.00                & 50.00±0.00                \\
30                     & 99.18 & 100.00   & 50.00±2.80              & 43.38±2.80           \\
31                     & 99.37 & 100.00   & 50.00±1.34             & 42.39±1.34          \\ \hline

\end{tabular}}}
\caption{Probing results for LLaMA-8B. Bolded values indicate statistical significance (p \textless 0.01).LDT denotes Linear Domain Transfer, and NLDT denotes Nonlinear Domain Transfer.}
\label{tab:8b_probing_results}

\end{table}

\clearpage
\section{Attention Maps} \label{sec:appendix:attention}
We will discuss the findings and highlights from both the self-attention maps and the cross-attention maps.
\paragraph{Self-Attention Maps}

\begin{figure*}[htbp]
    \centering
    \includegraphics[width=\textwidth]{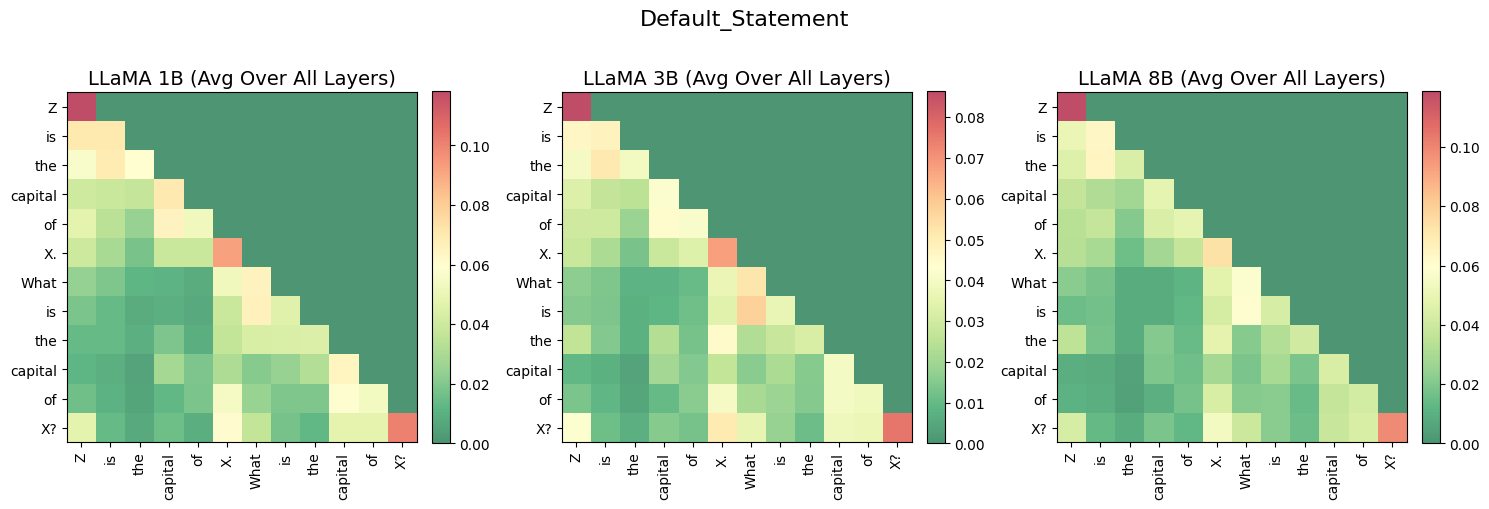}
    \caption{Self-attention maps between prompts and model responses across statement types of the World Capitals dataset from Llama3.}
    \label{fig:wc_sa_d}
\end{figure*}

\begin{figure*}[htbp]
    \centering
    \includegraphics[width=\textwidth]{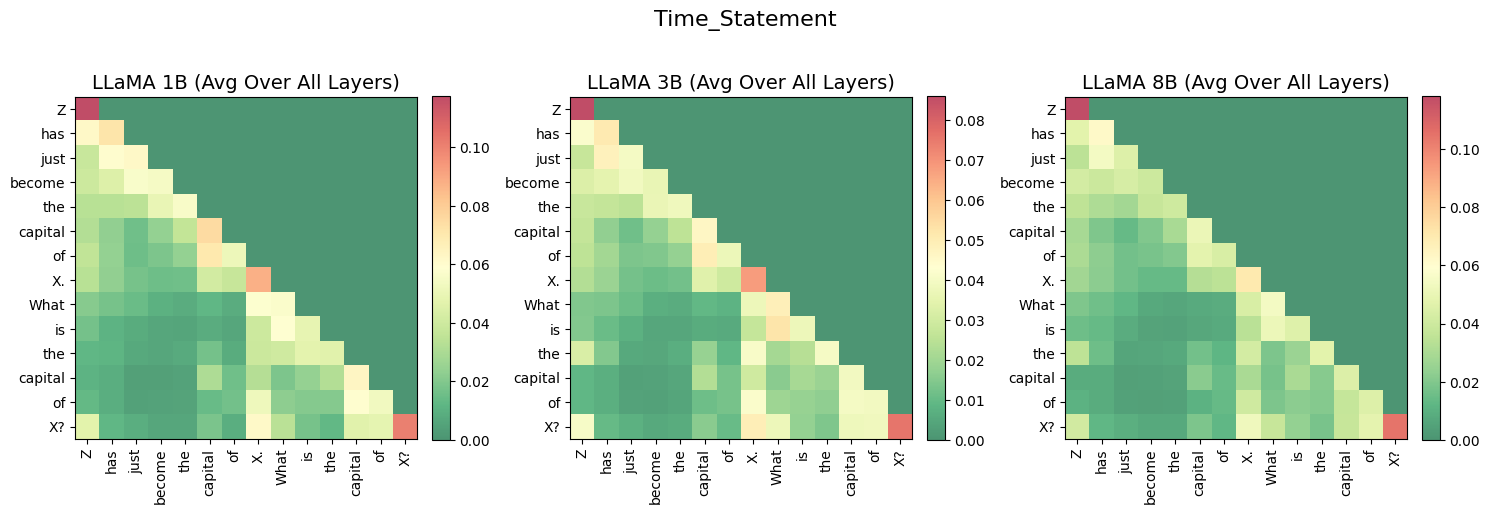}
    \caption{Self-attention maps between prompts and model responses across statement types of the World Capitals dataset from Llama3.}
    \label{fig:wc_sa_t}
\end{figure*}

\begin{figure*}[htbp]
    \centering
    \includegraphics[width=\textwidth]{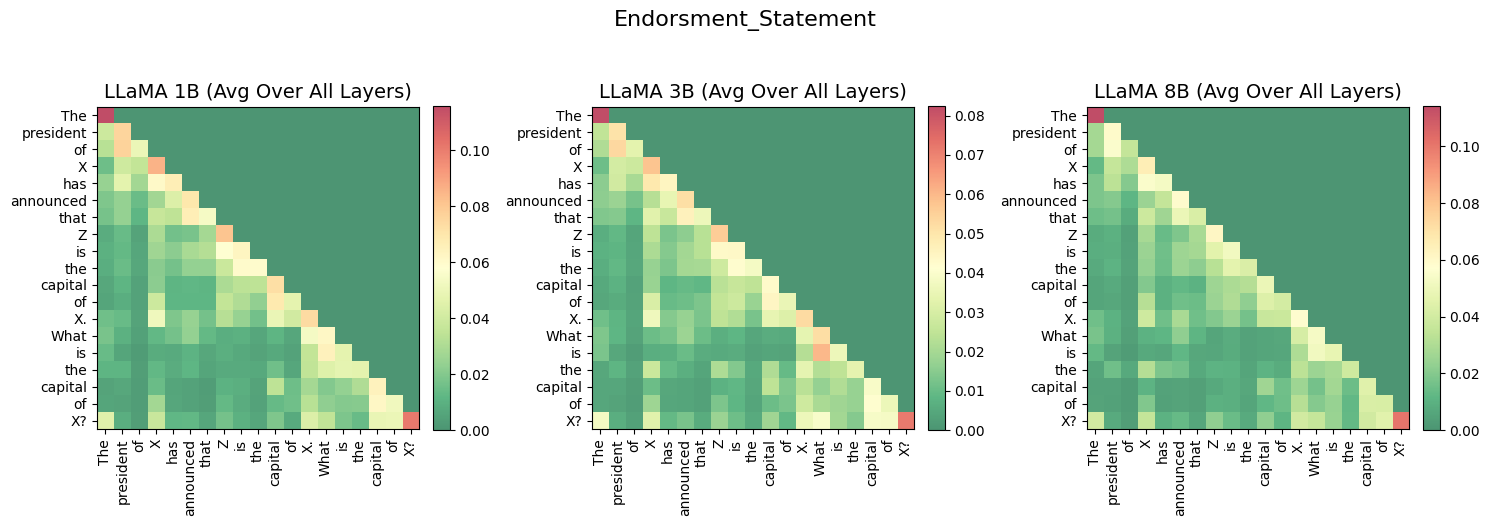}
    \caption{Self-attention maps between prompts and model responses across statement types of the World Capitals dataset from Llama3.}
    \label{fig:wc_sa_e}
\end{figure*}

\begin{figure*}[htbp]
    \centering
    \includegraphics[width=\textwidth]{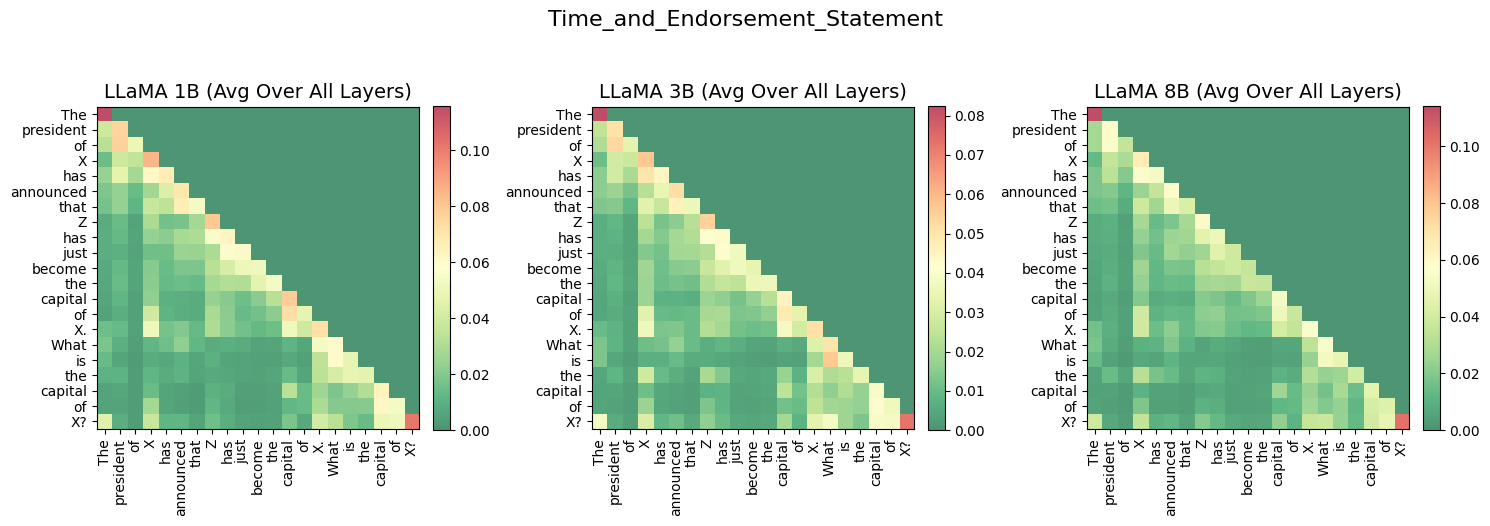}
    \caption{Self-attention maps between prompts and model responses across statement types of the World Capitals dataset from Llama3.}
    \label{fig:wc_sa_te}
\end{figure*}

Figures~\ref{fig:wc_sa_d}, \ref{fig:wc_sa_t}, \ref{fig:wc_sa_e}, and \ref{fig:wc_sa_te} illustrate the self-attention maps across various combinations of statement types and model sizes. Examining the default and time statement setups, as shown in Figures~\ref{fig:wc_sa_d} and \ref{fig:wc_sa_t}, reveals that the conflicting capital (Z) and the country (X) is successfully attended to, regardless of model size. However, for the endorsement and time+endorsement setups, depicted in Figures~\ref{fig:wc_sa_e} and \ref{fig:wc_sa_te}, respectively, all Llama models fail to attend to Z, especially for the 8B model. Instead, they predominantly focus on the \textit{The}. This pattern does not align with the results presented in Figure~\ref{fig:response_proportion:capital}, where the default statement exhibits the highest CK portion, while the time statement shows the lowest. Further investigation is necessary to understand this behavior.

A similar mismatch between the CK/PK portions and the self-attention maps is also observed in the Olympics Winners dataset. In this dataset, X, V, and Y represent the conflicting award receiver, verbal phrases (e.g., "is now," "will now be," and "has become," etc.), and temporal phrases (e.g., "yesterday," "today," and "last week," etc.), respectively. Although the LLMs tend to focus on incorrect words in this dataset, they still heavily rely on the conflicting knowledge. This observation leads us to conclude that self-attention maps do not provide significant insights into how LLMs emphasize certain information.

\begin{figure*}[htbp]
    \centering
    \includegraphics[width=\textwidth]{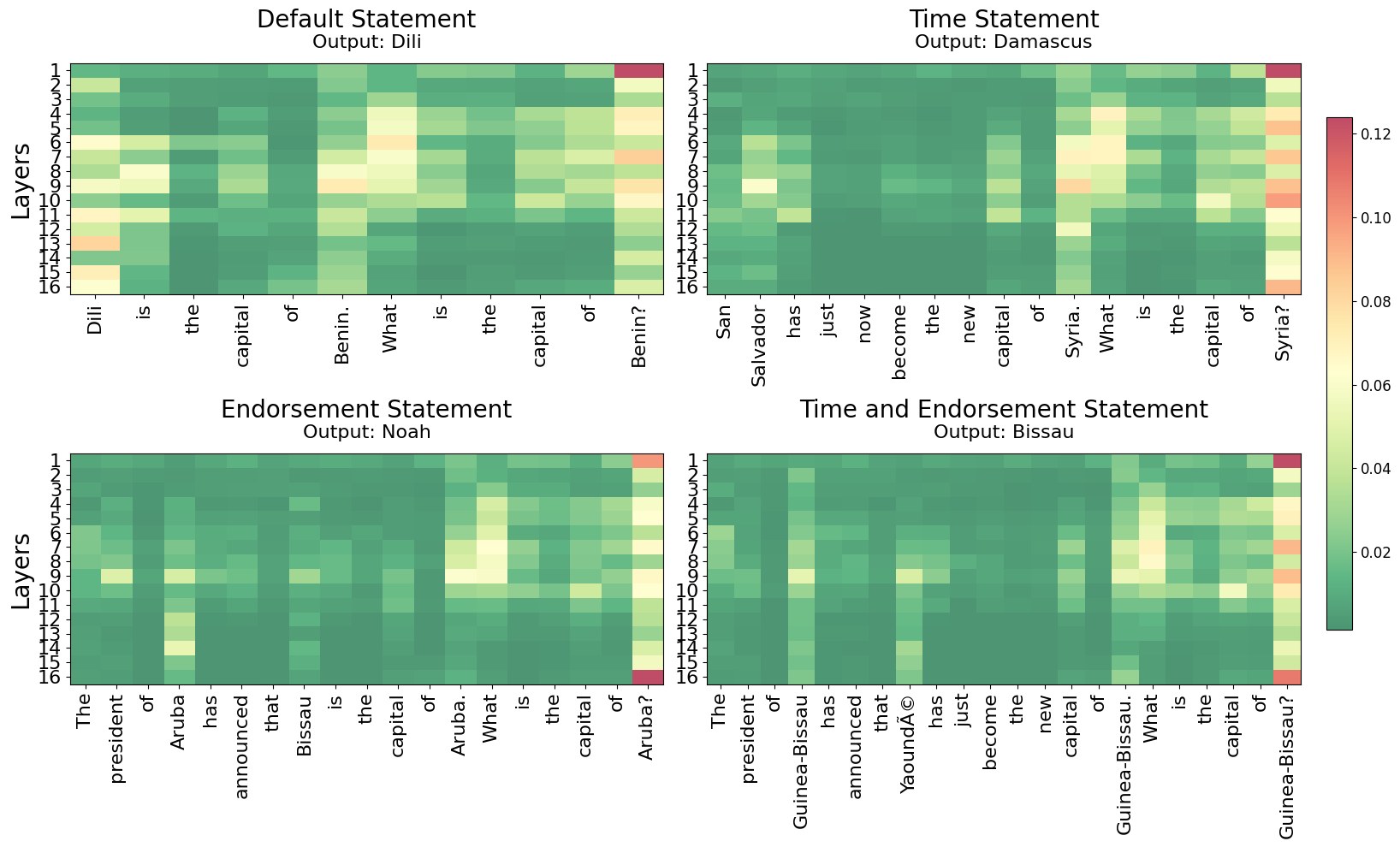}
    \caption{Cross-attention maps between prompts and model responses across statement types of the World Capitals dataset from Llama3 1B. Low attention values on statements indicate answers derived from the model’s parametric knowledge. Conversely, high attention values on the corresponding context highlight responses grounded in the provided context.}
    \label{fig:world_capitals_cross_attention}
\end{figure*}

\begin{figure*}[htbp]
    \centering
    \includegraphics[width=\textwidth]{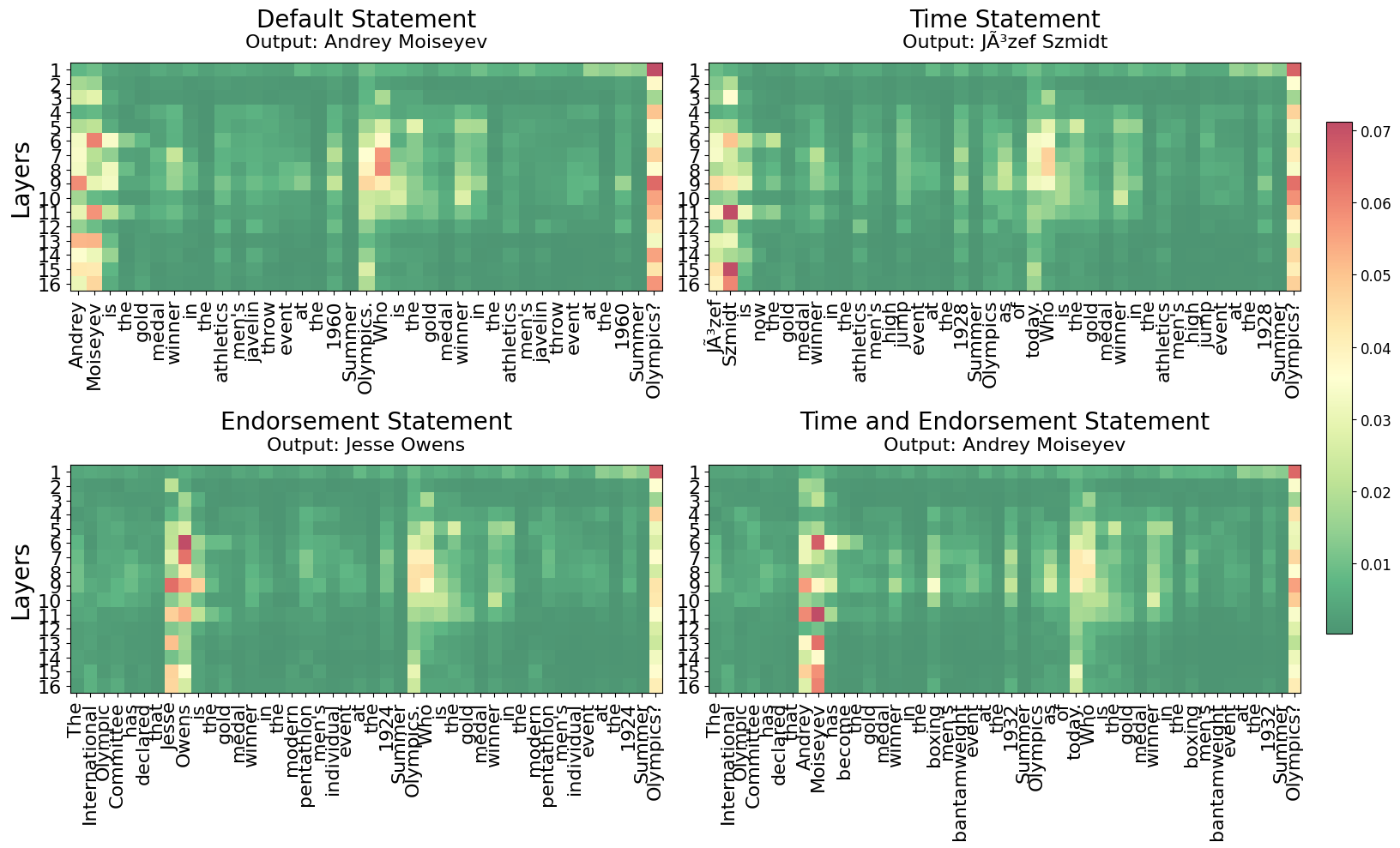}
    \caption{Cross-attention maps between prompts and model responses across statement types of the Olympics Winners dataset from Llama3 1B.}
    \label{fig:olympics_cross_attention}
\end{figure*}

\paragraph{Cross-Attention Maps}
We present cross-attention maps for the World Capitals and Olympics Winners datasets in Figure~\ref{fig:world_capitals_cross_attention} and Figure~\ref{fig:olympics_cross_attention}, respectively. These maps are derived from the max-pooled attention weights from each head at each layer. For clarity, we provide an example for each statement type: default, time, endorsement, and time and endorsement statements.

In Figure~\ref{fig:world_capitals_cross_attention}, under the default statement setup, Llama demonstrates increased attention to the conflicting capital, \textit{Dili}, across multiple layers. Notably, the attention on \textit{Dili} intensifies in deeper layers, which are closer to the final decision, indicating that the model’s response relies more on the provided context rather than its parametric knowledge. Conversely, an opposite trend is observed in the other three statement types. While some context is lightly attended, the attention values remain significantly stronger in the final layer, reflecting a greater reliance on the model’s parametric knowledge.

Figure~\ref{fig:olympics_cross_attention} presents the cross-attention maps for the Olympics Winners dataset. According to Table~\ref{table:olympic}, Llama3 1B's responses completely relies on the external context. Regardless of the question type, when the LM relies on contextual information to answer, it effectively attends to the relevant keywords required for generating the response using external knowledge.

\begin{figure*}[htbp]
    \centering
    \includegraphics[width=\textwidth]{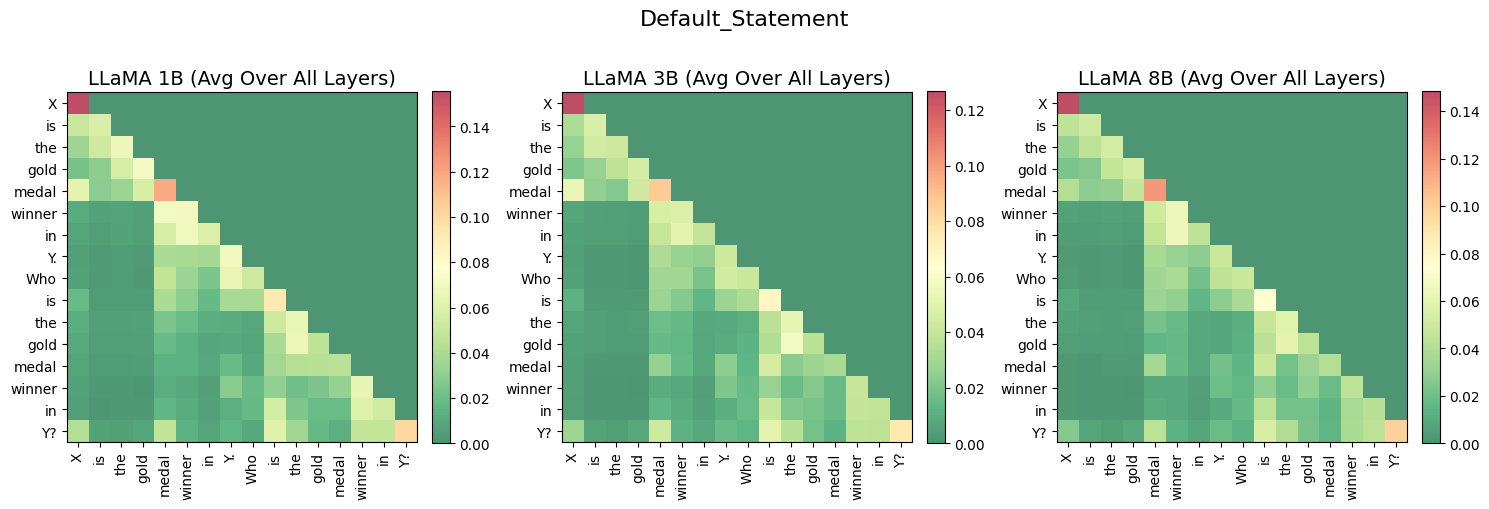}
    \caption{Self-attention maps between prompts and model responses across statement types of the Olympics Winners and World Capitals dataset from Llama3.}
    \label{fig:ow_sa_d}
\end{figure*}

\begin{figure*}[htbp]
    \centering
    \includegraphics[width=\textwidth]{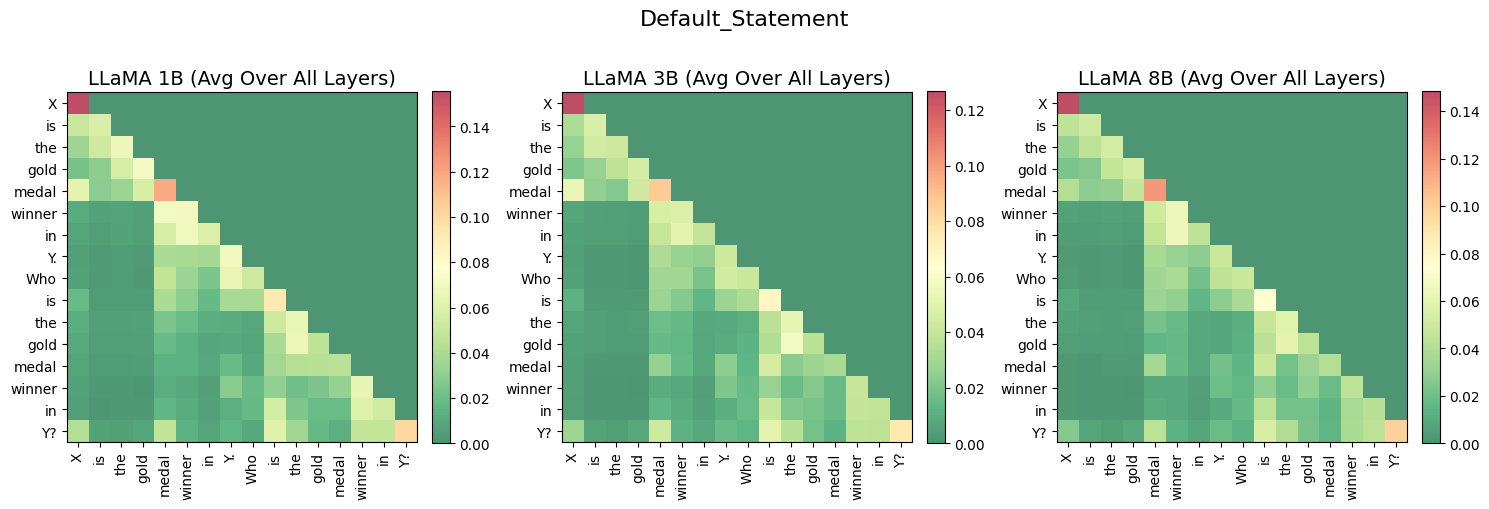}
    \caption{Self-attention maps between prompts and model responses across statement types of the Olympics Winners and World Capitals dataset from Llama3.}
    \label{fig:ow_sa_t}
\end{figure*}

\begin{figure*}[htbp]
    \centering
    \includegraphics[width=\textwidth]{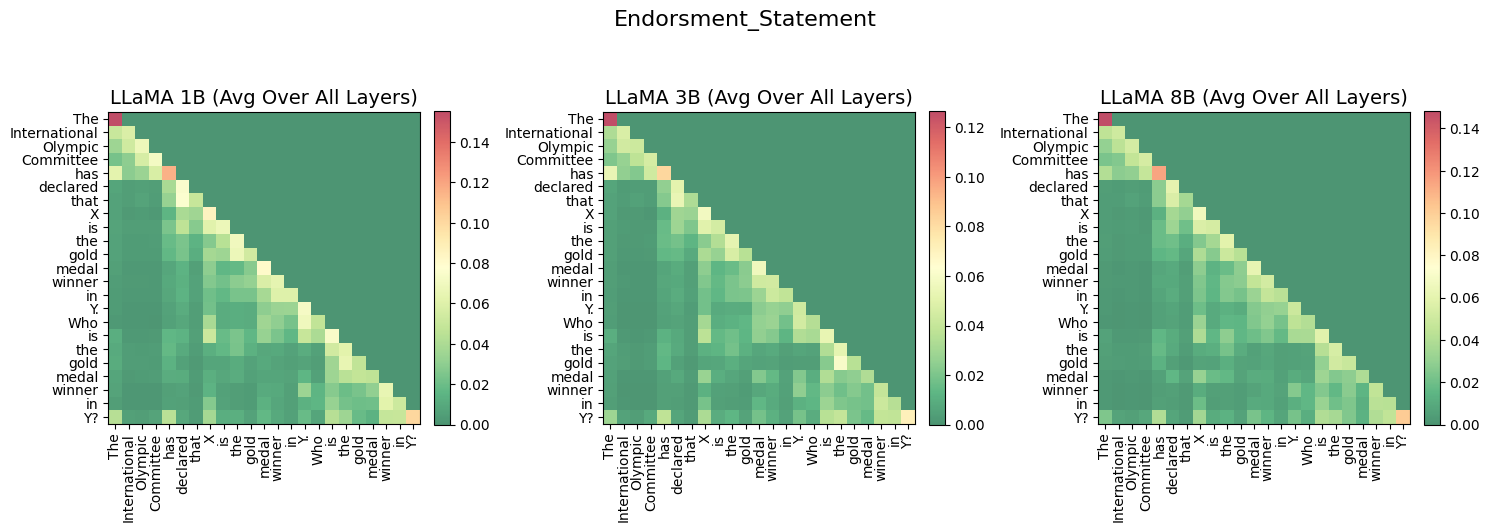}
    \caption{Self-attention maps between prompts and model responses across statement types of the Olympics Winners and World Capitals dataset from Llama3.}
    \label{fig:ow_sa_e}
\end{figure*}

\begin{figure*}[htbp]
    \centering
    \includegraphics[width=\textwidth]{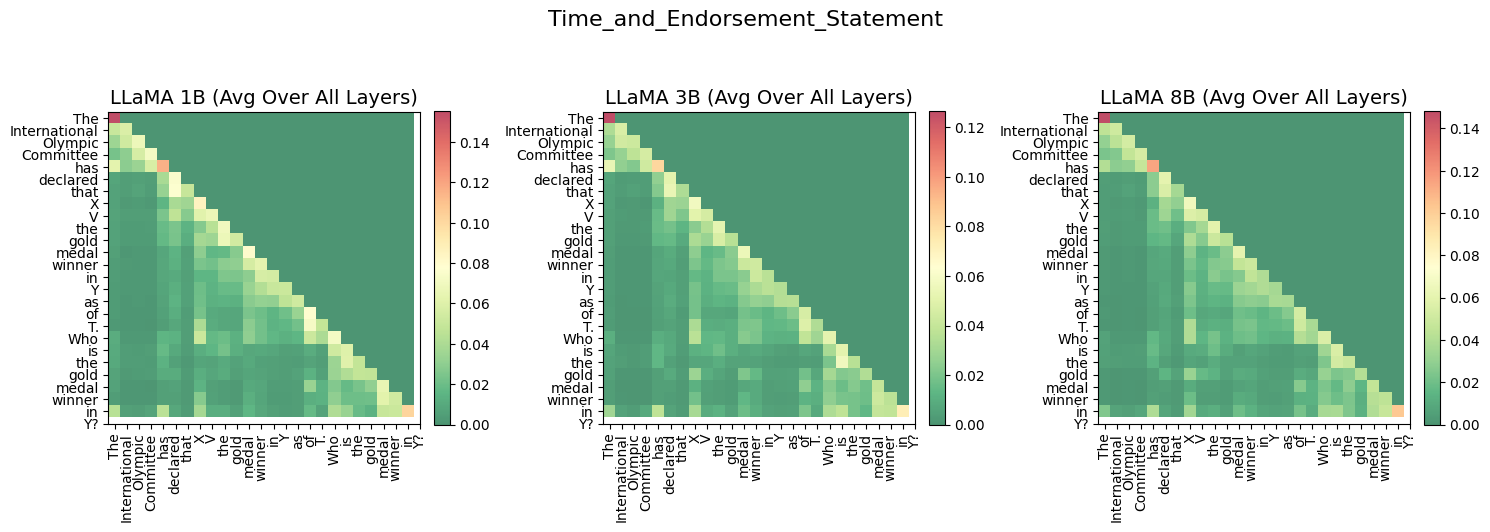}
    \caption{Self-attention maps between prompts and model responses across statement types of the Olympics Winners and World Capitals dataset from Llama3.}
    \label{fig:ow_sa_te}
\end{figure*}

\begin{figure*}[htbp]
    \centering
    \includegraphics[width=\textwidth]{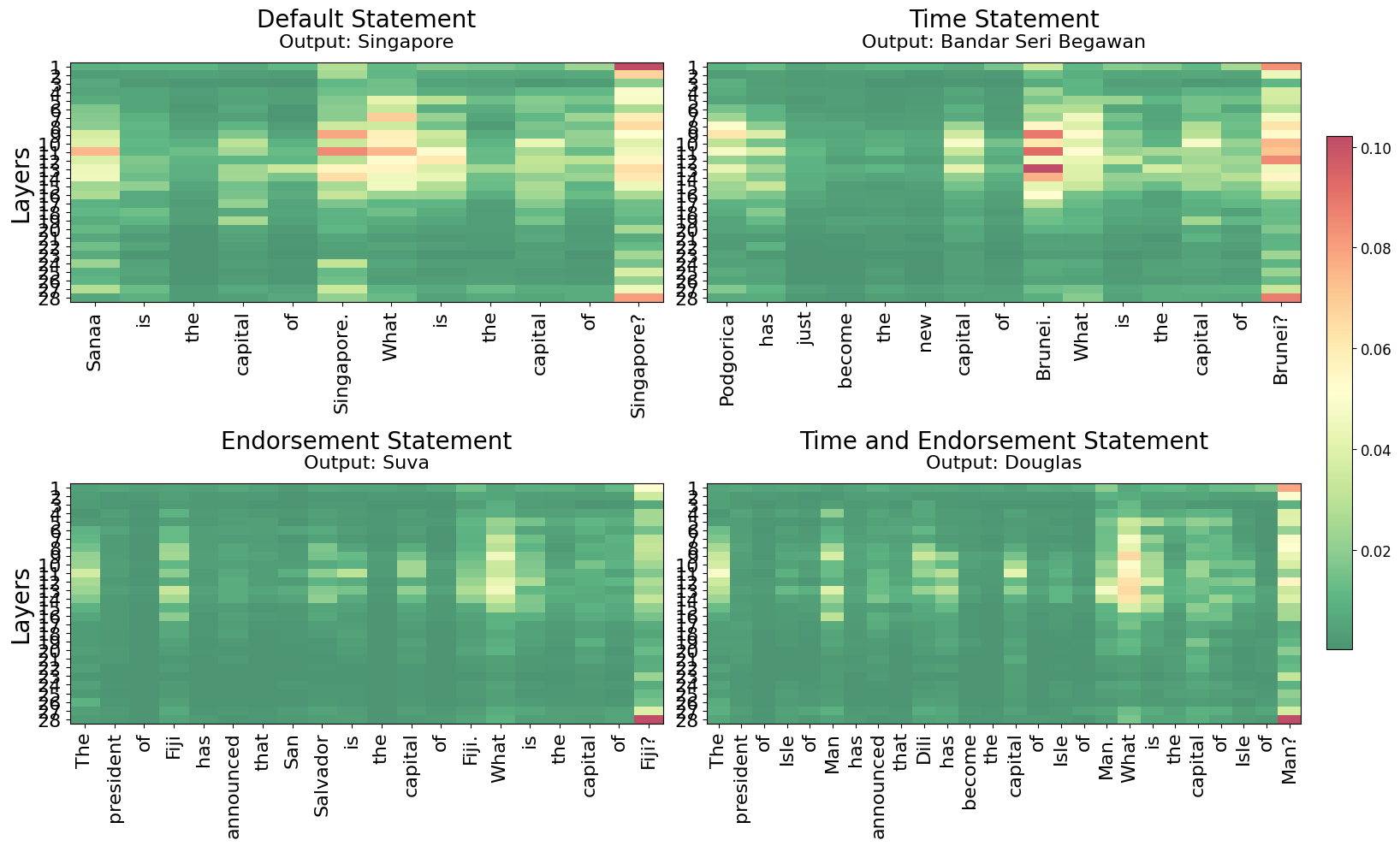}
    \caption{Cross-attention maps between prompts and model responses across statement types of the World Capitals dataset from Llama3 3B.}
    \label{fig:wc_ca_3b}
\end{figure*}

\begin{figure*}[htbp]
    \centering
    \includegraphics[width=\textwidth]{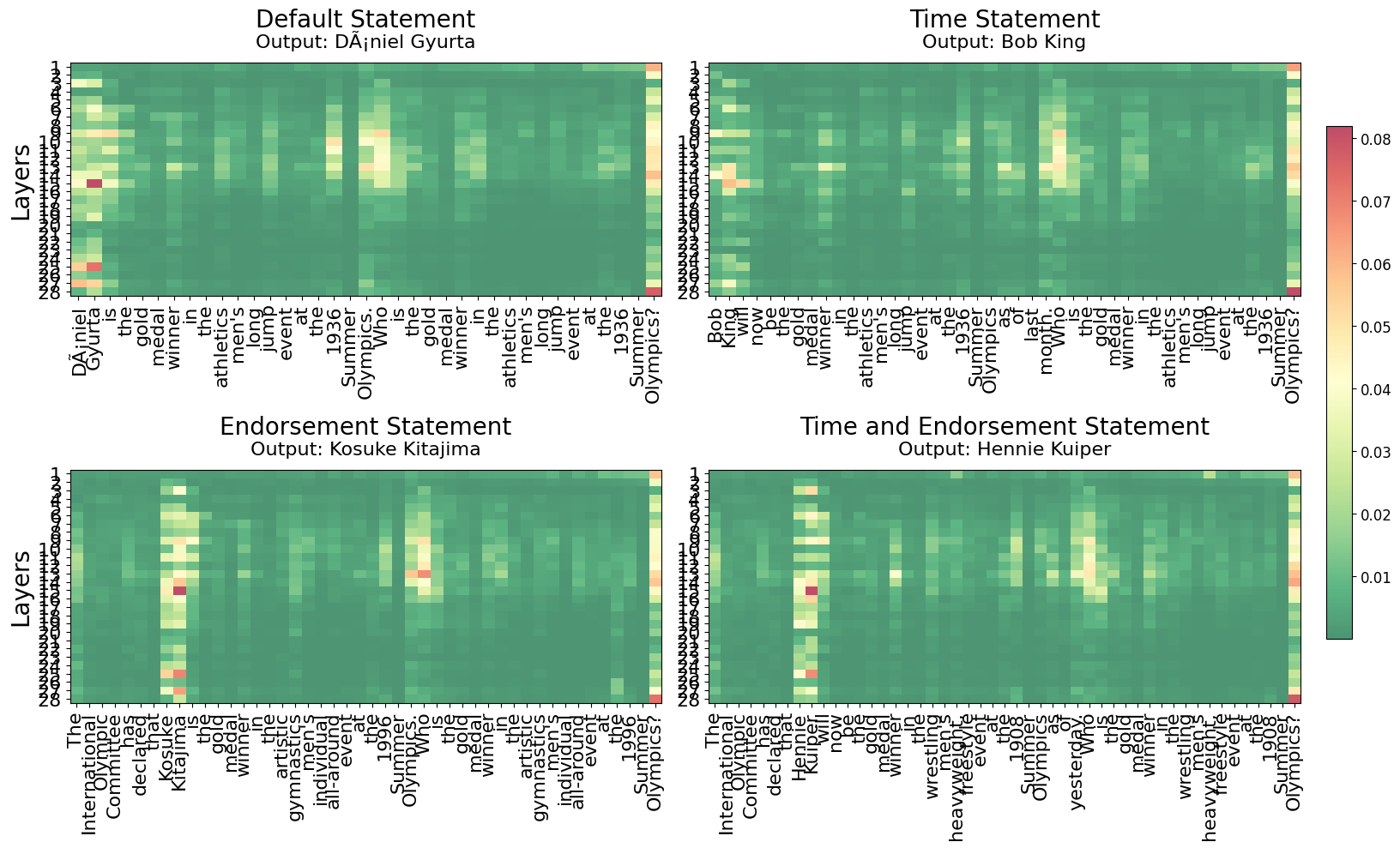}
    \caption{Cross-attention maps between prompts and model responses across statement types of the Olympics Winners dataset from Llama3 3B.}
    \label{fig:ow_ca_3b}
\end{figure*}

\begin{figure*}[htbp]
    \centering
    \includegraphics[width=\textwidth]{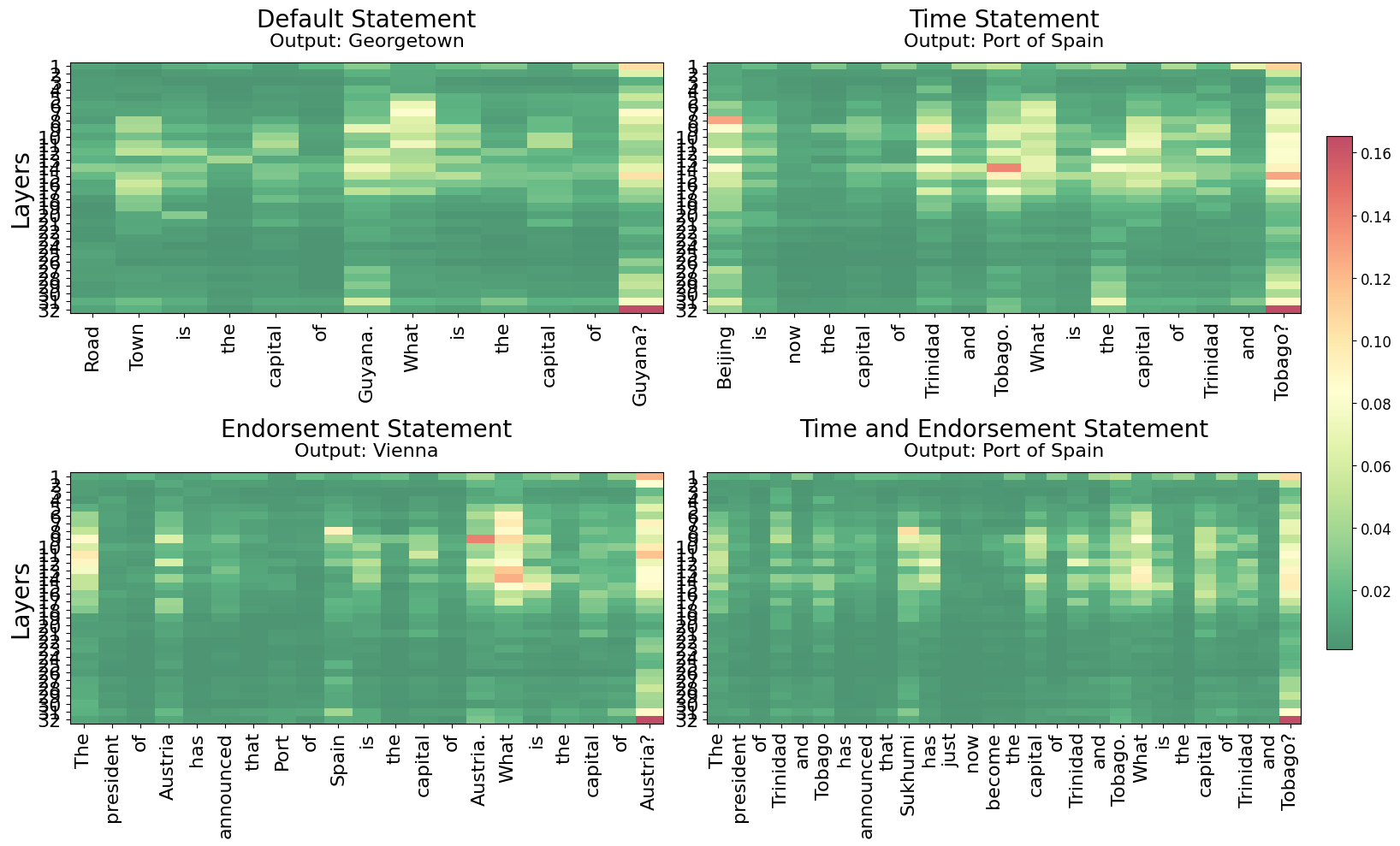}
    \caption{Cross-attention maps between prompts and model responses across statement types of the World Capitals dataset from Llama3 8B.}
    \label{fig:wc_ca_8b}
\end{figure*}

\begin{figure*}[htbp]
    \centering
    \includegraphics[width=\textwidth]{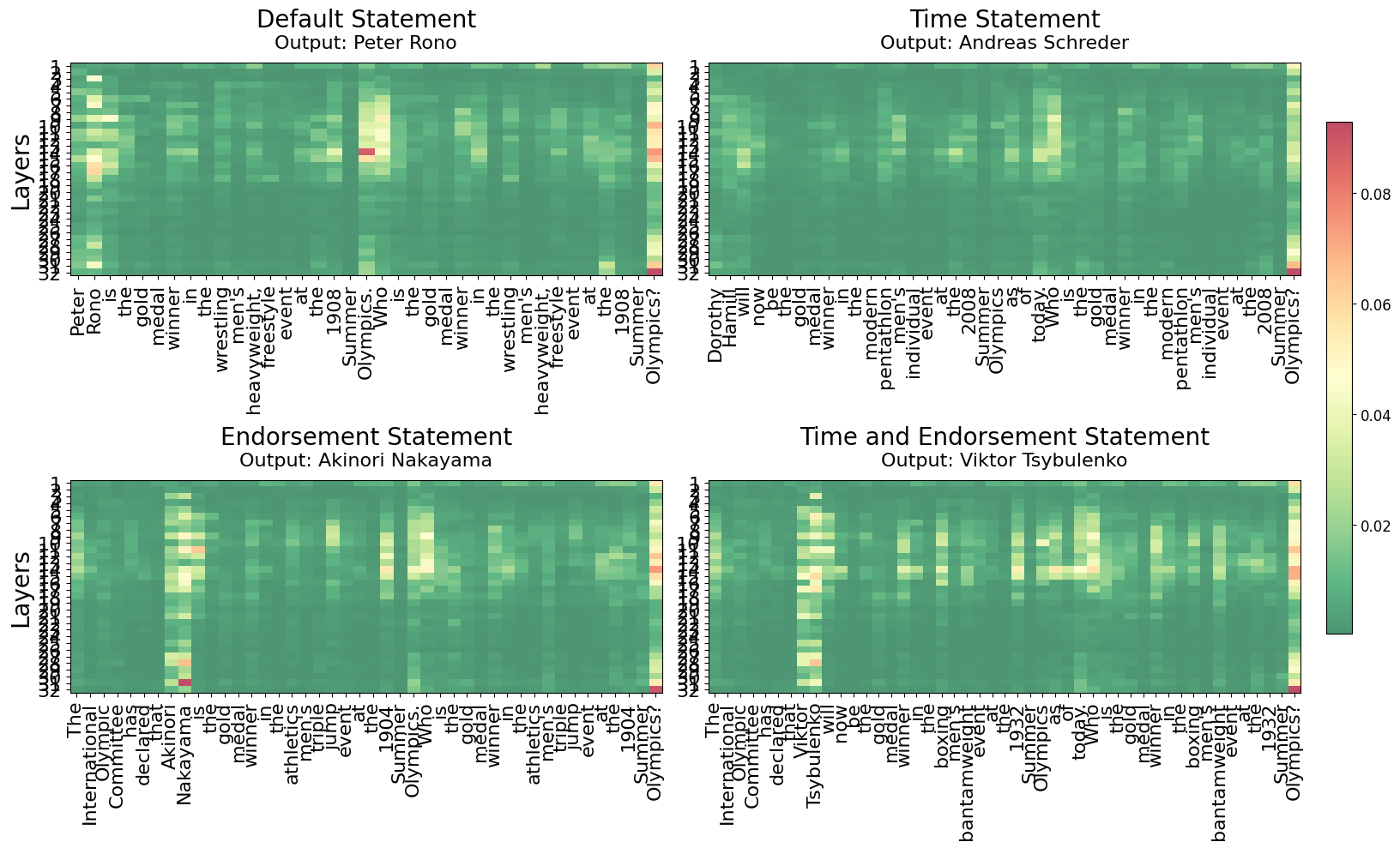}
    \caption{Cross-attention maps between prompts and model responses across statement types of the Olympics Winners dataset from Llama3 8B.}
    \label{fig:ow_ca_8b}
\end{figure*}


\end{document}